\documentclass{article}

    \usepackage[final]{neurips_2025}

\usepackage[utf8]{inputenc} % allow utf-8 input
\usepackage[T1]{fontenc}    % use 8-bit T1 fonts
\usepackage{hyperref}       % hyperlinks
\usepackage{url}            % simple URL typesetting
\usepackage{booktabs}       % professional-quality tables
\usepackage{amsfonts}       % blackboard math symbols
\usepackage{amssymb}        % for \Letter symbol
\usepackage{nicefrac}       % compact symbols for 1/2, etc.
\usepackage{microtype}      % microtypography
\usepackage[table]{xcolor}         % colors
\usepackage{todonotes}
\usepackage{amsmath}
\usepackage{cleveref}
\usepackage{tablefootnote}
\usepackage{makecell}
\usepackage[misc]{ifsym}
\crefname{section}{Sec.}{Secs.}
\crefname{equation}{Eq.}{Eqs.}
\crefname{table}{Tab.}{Tabs.}
\crefname{figure}{Fig.}{Figs.}

\title{TC-Light: Temporally Coherent Generative Rendering for Realistic World Transfer}

\author{
    Yang~Liu$^{1,2}$ \quad
    Chuanchen~Luo$^{3}$ \quad
    Zimo~Tang$^{6}$ \quad
    Yingyan~Li$^{1,2}$ \quad
    \textbf{Yuran~Yang}$^{5}$ \\[1pt]
    \textbf{Yuanyong~Ning}$^{5}$ \quad
    \textbf{Lue~Fan}$^{1,2}$ \quad
    \textbf{Junran~Peng}$^{4 *}$ \quad
    \textbf{Zhaoxiang~Zhang}$^{1,2}$\thanks{Correponding author.}\\[3pt]
    $^1$NLPR, MAIS, Institute of Automation, Chinese Academy of Sciences, \\
    $^2$University of Chinese Academy of Sciences \quad
    $^3$Shandong University \\
    $^4$University of Science and Technology Beijing \quad
    $^5$Tencent \\
    $^6$Huazhong University of Science and Technology\\[3pt]
    \small{\texttt{\{liuyang2022, liyingyan2021, lue.fan, zhaoxiang.zhang\}@ia.ac.cn}}\\
    \small{\texttt{u202315173@hust.edu.cn}} \quad 
    \small{\texttt{yangyuran@bupt.edu.cn}} \quad \small{\texttt{yyning@tencent.com}}\\
    \small{\texttt{chuanchen.luo@sdu.edu.cn}}
    \quad
    \small{\texttt{jrpeng4ever@126.com}}
}

\begin{document}

\maketitle

\begin{abstract}

Illumination and texture rerendering are critical dimensions for world-to-world transfer, which is valuable for applications including sim2real and real2real visual data scaling up for embodied AI. Existing techniques generatively re-render the input video to realize the transfer, such as video relighting models and conditioned world generation models. Nevertheless, these models are predominantly limited to the domain of training data (e.g., portrait)  or fall into the bottleneck of temporal consistency and computation efficiency, especially when the input video involves complex dynamics and long durations. In this paper, we propose \textbf{TC-Light}, a novel paradigm characterized
by the proposed two-stage post optimization mechanism. Starting from the video preliminarily relighted by an inflated video relighting model, it optimizes appearance embedding in the first stage to align global illumination. Then it optimizes the proposed canonical video representation, i.e., \textbf{Unique Video Tensor (UVT)}, to align fine-grained texture and lighting in the second stage. To comprehensively evaluate performance, we also establish a long and highly dynamic video benchmark. Extensive experiments show that our method enables physically plausible re-rendering results with superior temporal coherence and low computation cost. The code and video demos are available at our \href{https://dekuliutesla.github.io/tclight/}{Project Page}.
\end{abstract}

\section{Introduction}
\label{sec: intro}

Lighting and its interaction with both real and synthetic environments fundamentally shapes how humans—and embodied agents—perceive the world. The ability to re-render the illumination and texture (or so-called relighting) of captured image sequences, especially in complex, highly dynamic scenes, is critically valuable for various world-to-world transfer use cases like filmmaking \citep{Richardt_Stoll_Dodgson_Seidel_Theobalt_2012} and augmented reality \citep{li2022physically}. Crucially, by re-rendering the CG-simulated or realistic video data used to train embodied agents, it can bridge the sim-to-real gap and enable real-to-real transfer, thus unlocking access to massive high-quality data that is essential for stepping towards embodied intelligence.

\begin{figure}
    \centering
    \includegraphics[width=0.99\linewidth]{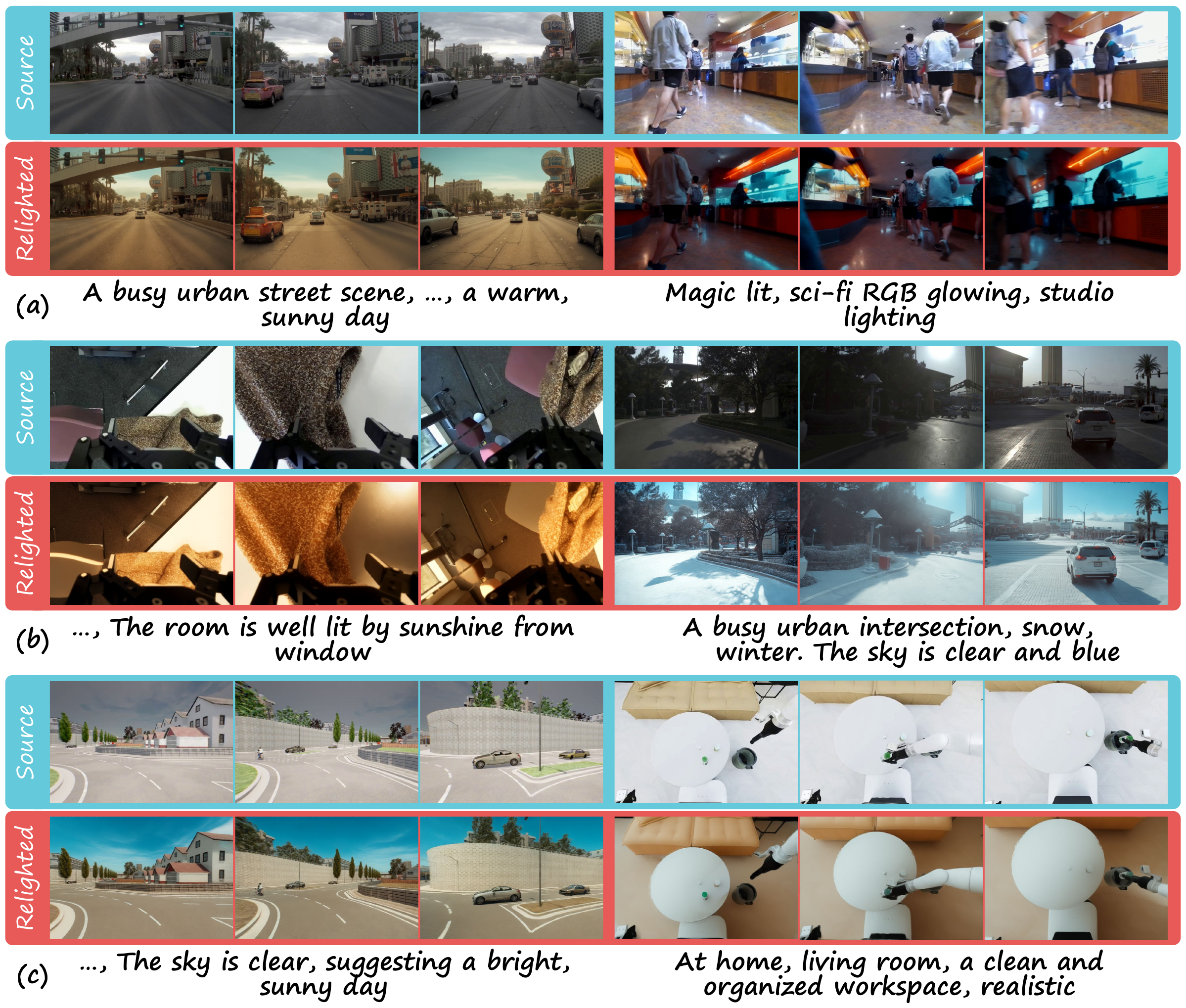}
    \caption{Relighting results on long videos under various dynamic scenes, averaging 256 frames per clip. Though the video involves frequent changes of foreground objects (row \textbf{(a)}), highly dynamic camera motions (row \textbf{(b)}), the TC-Light realizes consistent and physically plausible relighting results. Row \textbf{(c)} also shows its potential to mitigate the sim2real gap for synthetic renderings.}
    \label{fig: teaser}
\end{figure}

Despite its importance, the video illumination and texture rerendering remains a highly challenging problem, particularly when \textbf{camera motion is highly dynamic} and \textbf{foreground objects frequently enter and exit scenes}, as shown in \cref{fig: teaser}. Most existing generative relighting techniques \cite{zhang2025scaling, li2022physically, jin2024neural_gaffer, ren2024relightful, kim2024switchlight, wang2023sunstage} are tailored for static images. As shown in \cref{subsec: comparison}, naively inflating them to a video model with existing zero-shot strategies struggles to balance the consistency and quality. Moreover, the considerable training cost and scarcity of video lighting datasets hinder fine-tuning a pretrained model for this task. Besides, though generative video relighting and world generation models are emerging, they are either restricted on domain of training data \cite{zhang2021neural, choi2024personalized, cai2024real, alhaija2025cosmos} or burdened by considerable computation overhead \cite{zhou2025light, fang2025relightvid} on long video, as validated in \cref{subsec: comparison}. 

To address the limitations outlined above, we propose \textbf{TC-Light}. We utilize the SOTA image relighting model IC-Light \cite{zhang2025scaling} as the baseline, and inflate it to a video model in a zero-shot manner with our decayed multi-axis denoising model, which is distinguished by the proposed Decayed Noise Weighting and Noise Statistic Alignment. It provides a preliminary video relighting result. The core innovation of TC-Light lies in a two-stage post-optimization framework that substantially improves temporal consistency. The first stage introduces per-frame appearance embedding to compensate for exposure discrepancy. It is optimized with photometric loss against the preliminarily relighted video and a flow-based loss between adjacent frames. This enforces global illumination consistency and facilitates consequent optimization. The second stage compresses the output to a canonical representation, i.e., \textbf{Unique Video Tensor (UVT)}, according to priors including optical flow and depth of the source video. UVT is then optimized by minimizing the warping error across decompressed frames while aligning the content with the first stage result. As shown in \cref{tab: comparison}, our optimization procedure is extremely efficient and introduces minimal VRAM overhead.

To comprehensively assess the effectiveness of our model, we introduce a challenging benchmark tailored for complex and highly dynamic scenes. It comprises 58 videos of averagely 256 frames per clip, spanning both indoor and outdoor environments, realistic and synthetic settings, and a wide range of lighting and weather conditions. Extensive experiments demonstrate that our method achieves high-quality, temporally consistent video relighting while maintaining low computational overhead, highlighting its great potential for downstream applications such as embodied AI. Our main contributions are as follows:

\begin{itemize}

\item[$\bullet$] A novel optimization-based
video relighting paradigm for long videos with high and complicated dynamics, significantly improving the temporal consistency of the relighting result.

\item[$\bullet$] We establish a new long-video relighting benchmark characterized by high motion dynamics and broad scene diversity, covering various environments and data domains. 

\item[$\bullet$] Extensive experiments validate that our method achieves SOTA performance in producing temporally consistent, naturally relighted videos with minimal computational cost.

\end{itemize}

\section{Related Work}
\label{sec: related works}

\subsection{Learning-based Illumination Editing}
\label{subsec: illumination}
Over the past few years, deep neural networks have become one of the main forces behind research in the field of illumination control. Pioneering works \cite{sun2019single, nestmeyer2020learning, pandey2021total, das2021dsrn} train convolutional encoder–decoder networks on light-stage data. The learned prior knowledge enables models to relight a portrait according to the specified light conditions. More recently, large diffusion-based generators have gained popularity for illumination editing. LightIt \cite{kocsis2024lightit} explicitly conditions the diffusion process on estimated shading and normal maps, giving fine-grained lighting control ability, while SwitchLight \cite{kim2024switchlight} incorporates a physics-guided architecture to simulate light-surface interactions better. \cite{zhang2024lumisculptconsistencylightingcontrol, bharadwaj2024genlitreformulatingsingleimagerelighting} leverage video foundation models to generate realistic lighting variations over a static image. IC-Light \cite{zhang2025scaling}, the current state of the art, learns illumination mixture and decomposition from a large quantity of data. Building on these advances in image relighting, video relighting has started to gain traction. \cite{zhang2021neural, choi2024personalized} learns to disentangle light and intrinsic appearance on portrait videos. \cite{cai2024real} represents talking faces as relightable NeRFs guided by predicted albedo and shading features. Extending IC-Light, Light-A-Video \cite{zhou2025light} introduces zero-shot cross-frame attention modification, while RelightVid \cite{fang2025relightvid} trains a temporally inflated IC-Light with a carefully designed video relighting dataset. However, these methods are either restricted to portrait scenarios or struggle with computational efficiency on long videos. In contrast, our model delivers high-quality relighting with strong temporal consistency and low computation cost, even in complex and highly dynamic scenes.

\subsection{Diffusion-based Video Editing}
\label{subsec: video editing}
The diffusion model \cite{ho2020denoising} has become the go-to model for visual domain transfer and content editing. Based on training paradigms, recent advancements can be grouped into three categories: \textbf{(i) training-based} models extend pretrained image diffusion models with temporal layers and are trained on large-scale video datasets, such as \cite{liew2023magicedit, chen2023controlavideo, molad2023dreamixvideodiffusionmodels, ma2024maskint, qin2024instructvid2vid, wang2025videodirector}. CCEdit \cite{feng2024ccedit} and FlowVid \cite{liang2024flowvid} further integrate depth and flow cues for improved consistency and control. \textbf{(ii) training-free} models mainly rely on cross-frame attention to enforce temporal coherence. TokenFlow \cite{tokenflow2023} and FLATTEN \cite{cong2023flatten} guide attention using estimated optical flow. RAVE \cite{kara2024rave} enhances latent interactions by denoising over a reorganized latent grid, while Slicedit \cite{cohen2024slicedit} uses spatiotemporal slices to inject motion priors. VidToMe \cite{li2024vidtome}, on the other hand, exploits temporal redundancy through token merging and unmerging. \textbf{(iii) one-shot-tuned} models typically learn a canonical video representation in a few iterations and propagate its edits across frames. StableVideo \cite{chai2023stablevideo} learns to represent video as a foreground and background atlas. CoDeF \cite{ouyang2024codef} learns a hash table and decoding MLP to map frames to a single canonical image. Video-3DGS \cite{shin2024enhancing} adapts deformable 3DGS \cite{kerbl3Dgaussians} to model input video. Our method combines \textbf{(ii)} and \textbf{(iii)} and proposes an explicit, compact, and efficient canonical representation, i.e., Unique Video Tensor. It enables optimization to be finished within several minutes, which is much faster than 10-30 minutes cost \cite{shin2024enhancing} of CoDF and Video-3DGS. Our method also inherits the diffusion model design from training-free algorithms to reduce overall memory and time cost, enabling the processing of long videos.

\section{Method}
\label{sec: method}

In this section, we first introduce the task setting and preliminary knowledge about latent diffusion models in \cref{subsec: preliminary}. \cref{subsec: video model} further illustrates how our proposed Decayed Noise Weighting and Noise Statistics Alignment helps effectively lift the image relighting model to video space. \cref{subsec: post processing} details how our \textbf{key innovation}, i.e., the two-stage temporal consistency optimization strategy, helps align overall illumination and texture appearance.

\subsection{Preliminaries}
\label{subsec: preliminary}

\begin{figure}
    \centering
    \includegraphics[width=0.99\linewidth]{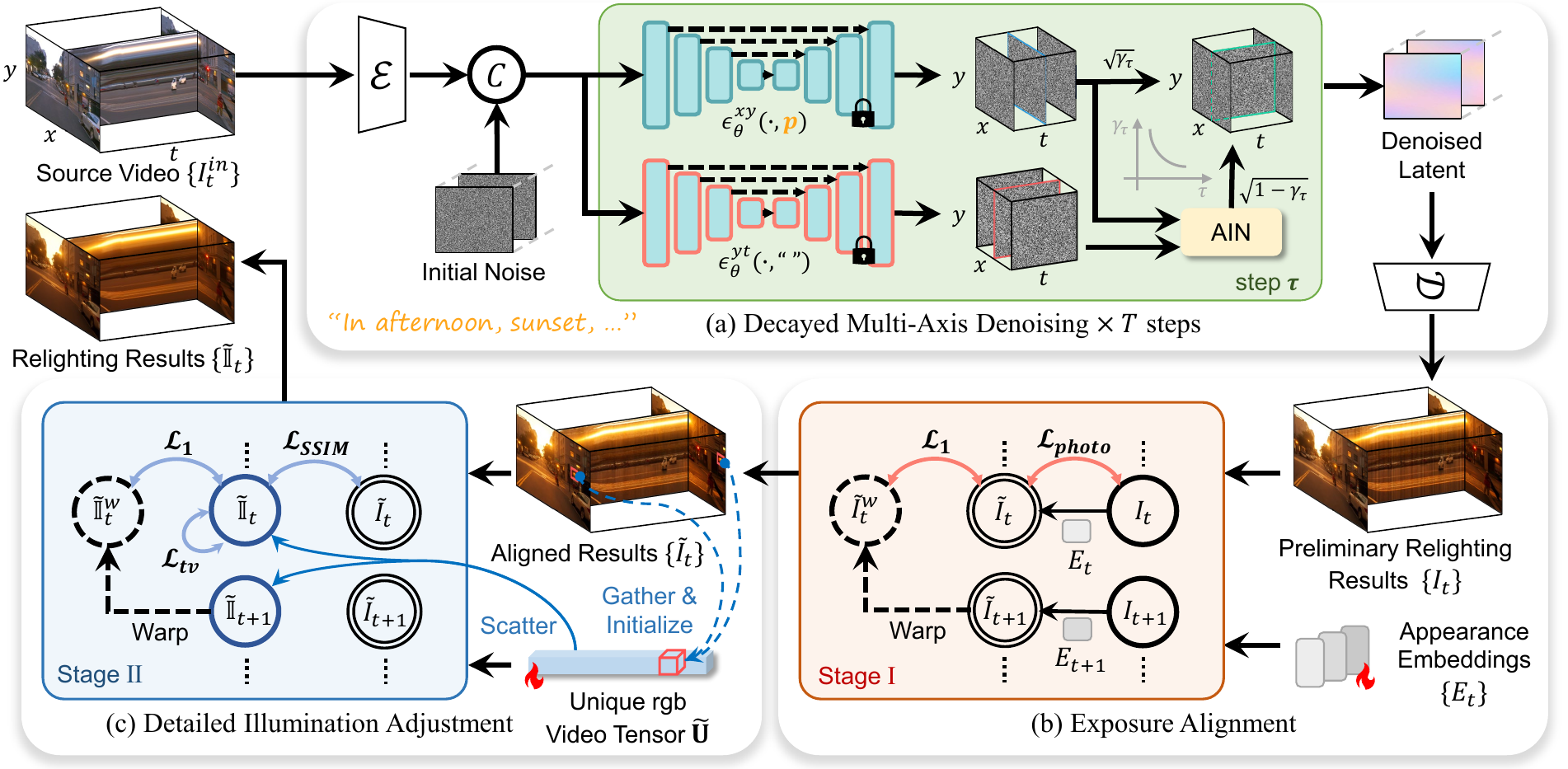}
    \caption{TC-Light overview. Given the source video and text prompt $p$, the model tokenizes input latents in $xy$ plane and $yt$ plane separately. The predicted noises are adaptively combined together for denoising (cf. \cref{subsec: video model}). Its output then undergoes two-stage optimization to enhance temporal consistency of illumination and texture, which are respectively detailed in \cref{subsubsec: exposure} and \cref{subsubsec: uvt}.}
    \label{fig: pipeline}
\end{figure}

\textbf{Task Setting.} As shown in \cref{fig: pipeline}, we take RGB video as input. The axes of the video space-time volume are denoted by $(x, y, t)$, where $xy$ planes correspond to video frames and $yt$ planes are defined as \textbf{spatiotemporal slices} \cite{cohen2024slicedit}. Since the camera motion is highly dynamic, the target illumination can no longer be simply appointed by a static image or an HDR environment map. 
Due to superior flexibility and operability, 
we use textual prompts as the control signal and relight the entire frame.

\textbf{Latent Diffusion Models (LDMs)}. Denoising Diffusion Probabilistic Models (DDPMs) \cite{ho2020denoising} are a class of generative models that aim to recover target data distribution through an iterative denoising process. Due to the high computational cost of operating directly in pixel space, LDMs \cite{ramesh2022hierarchical, rombach2022high, saharia2022photorealistic} perform diffusion in a lower-dimensional latent space. Given a clean image $x_0$, and a pretrained autoencoder $\{\mathcal{E}(\cdot), \mathcal{D}(\cdot)\}$, LDMs first encode the image into latent space $z_0 = \mathcal{E}(x_0)$. The forward diffusion process then gradually  corrupts $z_0$ with Gaussian noise $\epsilon$ over time steps $\tau=1,...,T$

\begin{equation}
\label{eq: denoise}
    \begin{aligned}
        z_\tau=\sqrt{\alpha_\tau}z_0 + \sqrt{1-\alpha_\tau}\epsilon,
    \end{aligned}
\end{equation}

where $\{\alpha_\tau\}$ is a monotonically decreasing noise schedule. The reverse process begins from pure noise $z_T\sim\mathcal{N}(0, \textbf{\textit{I}})$. With guidance from control signal (image, text, depth, etc) $c$, the trained UNet \cite{ronneberger2015u} $\epsilon_\theta$ estimates the noise direction and progressively removes the noise from $z_T$. After the final denoising step, the estimated clean latent $\hat{z_0}$ is decoded by $ \mathcal{D}(\cdot)$ to obtain the generated image $\hat{x_0}=\mathcal{D}(\hat{z_0})$, which approximates the training distribution.

% Considering Sim2Real of visual signal mainly involves adjustment of texture distribution and light transportation, we base our experiment on IC-Light \cite{zhang2025scaling}, a large-scale image relighting diffusion model. 

% Besides, when the embodied agent executes tasks, it typically experiences an extremely dynamic change of viewpoint, and frequent in and out of human or other agents in the field of view. As a result, it is difficult to use background videos or HDR maps 

\subsection{Lifting Image Diffusion Model to Video Space}
\label{subsec: video model}

Considering outstand ability in physical plausibility and intrinsic property preservation, we adapt IC-Light \cite{zhang2025scaling} into a zero-shot video diffusion model. Concretely, we \textbf{(i)} enhance its diffusion blocks to capture spatiotemporal dependencies and \textbf{(ii)} introduce consistency prior from original frames. For \textbf{(i)}, we apply the token merging and unmerging technique of VidToMe \cite{li2024vidtome} to self-attention blocks. It divides the video frames into chunks and applies intra-chunk local token merging and inter-chunk global token merging, enabling short- and long-term consistency. The derived model $\epsilon_\theta$ serves as the basis of \textbf{(ii)}. Since it reduces the token count fed to the self-attention module, the computation cost is significantly decreased. For full details, please refer to the original VidToMe paper \cite{li2024vidtome}. 

% However, the VidToMe-style extension alone has limited enhancement in temporal consistency, and it still suffers illusions in the textureless area inherited from IC-Light.

For \textbf{(ii)}, we apply multi-axis denoising and adapt it with our Decayed Noise Weighting and Noise Statistics Alignment strategy. This modified version is named \textbf{decayed multi-axis denoising}. Similar to Slicedit \cite{cohen2024slicedit}, the denoiser has two components with shared weights
$\epsilon^{xy}_\theta(\cdot, p)$ that tokenizes each frame and merges tokens from local temporal slots, while $\epsilon^{yt}_\theta(\cdot, ``\,")$ tokenizes the $yt$ planes (cf. \cref{subsec: preliminary}) and merges tokens from local image width slot. Note that $\epsilon^{xy}_\theta$ conditions on target prompt $p$, while $\epsilon^{yt}_\theta$ takes empty prompt $``\,"$ as input (making the denoiser unconditional). The noises separately predicted by two parts according to the same input latents are combined together \cite{cohen2024slicedit}

% $\epsilon^{xy}_\theta(\cdot, p)$ that operates in image plane and chunks in time dimension, as well as $\epsilon^{yt}_\theta(\cdot, ``\,")$ that operates in $y-t$ slice and chunks in width dimension, as shown in part (a) of \cref{fig: pipeline}. Note that $\epsilon^{xy}_\theta$ conditions on target prompt $p$, while $\epsilon^{yt}_\theta$ takes empty prompt $``\,"$ as input (making the denoiser unconditional). The combined video denoiser of \cite{cohen2024slicedit} is:

\begin{equation}
\label{eq: slicedit}
    \begin{aligned}
        \epsilon^{V}_\theta(\cdot, p)=\sqrt{\gamma}\epsilon^{xy}_\theta(\cdot, p) + \sqrt{1-\gamma}\epsilon^{yt}_\theta(\cdot, ``\,"),
    \end{aligned}
\end{equation}

where hyperparameter $\gamma \in [0,1]$ balances effect from $\epsilon^{yt}_\theta$. However, the unconditional $\epsilon^{yt}_\theta$ would overly biases texture and lighting toward the source video, and therefore lead to unnatural relighting results, as validated in \cref{fig: vis} and \cref{fig: ablation}. To alleviate this problem, we introduce Decayed Noise Weighting, which replaces $\gamma$ with a timestep-dependent $\gamma_\tau$ that exponentially decays during denoising. To further align predicted noise from $\epsilon^{yt}_\theta$ to that of $\epsilon^{xy}_\theta$, we use Adaptive Instance Normalization (\textbf{AIN}) \cite{huang2017arbitrary} to align noise statistics

\begin{equation}
    \label{eq: slicedit-modified}
    \begin{aligned}
    \epsilon _{\theta}^{V}(\cdot ,p)=\sqrt{\gamma _{\tau}}\epsilon _{\theta}^{xy}(\cdot ,p)+\sqrt{1-\gamma _{\tau}}\widehat{\epsilon _{\theta}^{yt}}(\cdot ,``\,"),
    \end{aligned}
\end{equation}
\begin{equation}
    \label{eq: ain}
    \begin{aligned}
    \widehat{\epsilon _{\theta}^{yt}}(\cdot ,``\,")=\sigma _{\epsilon _{\theta}^{xy}}\left( \frac{\epsilon _{\theta}^{yt}(\cdot ,``\,")-\mu _{\epsilon _{\theta}^{yt}}}{\sigma _{\epsilon _{\theta}^{yt}}} \right) +\mu _{\epsilon _{\theta}^{xy}},
    \end{aligned}
\end{equation}

where $\mu_*$ and $\sigma_*$ are the channel-wise mean and standard deviation of each frame. This design preserves motion guidance from the source video while reducing unwanted texture and lighting bias, as validated by ablation studies in \cref{subsec: ablation}. The output denoised video is denoted as $\{I_t\}$.

\subsection{Post Optimization for Temporal Consistency}
\label{subsec: post processing}

Although the video diffusion extension in \cref{subsec: video model} has introduced spatial-temporal awareness and motion prior from the source video, noticeable illumination and texture flicker persist. To efficiently remove these artifacts, we introduce a two-stage post-optimization framework, as illustrated in parts (b) and (c) of \cref{fig: pipeline}.

\subsubsection{Stage I: Exposure Alignment}
\label{subsubsec: exposure}

As shown in part (b) of \cref{fig: pipeline}, the first stage introduces a per-frame appearance embedding $E_t$ to compensate for exposure misalignment between adjacent frames. Inspired by \cite{kerbl2024hierarchical}, we model $E_t$ as a $3\times4$ affine transformation matrix, initialized to the identity and optimized via Adam \cite{2015-kingma}.  Its supervision combines a photometric term with a flow-warp alignment term using hyperparameter $\lambda_e$

\begin{equation}
    \label{eq: exposure-loss}
    \begin{aligned}
    \mathcal{L} _{exposure}&=\left( 1-\lambda_e \right) \mathcal{L} _{photo}\left( \tilde{I}_t,I_t \right) +\lambda_e \mathcal{L} _1\left( \tilde{I}_t\odot M_t,\mathrm{Warp}_{t+1\rightarrow t}\left( \tilde{I}_{t+1} \right)\odot M_t \right) ,
    \end{aligned}
\end{equation}

where the homogeneously transformed pixel color $\tilde{I}_t\left( x,y \right) =E_t\left[ I_t\left( x,y \right) |1 \right] ^T$. The photometric loss $\mathcal{L} _{photo}$ is the weighted sum of L1 loss and D-SSIM loss \cite{kerbl2024hierarchical}, ensuring the transformed frame retains its original content and structure. The second term warps the next frame back to the current timestamp $t$, according to forward and backward flows $F_{fwd,t}$ and $F_{bwd,t}$ estimated through MemFlow \cite{dong2024memflow} or provided by the dataset. Then it applies an L1 penalty $\mathcal{L} _1$ to align their exposures. To mask out regions with unreliable flow or occlusion, we apply a soft mask $M_t$

\begin{equation}
    \label{eq: soft-mask}
    \begin{aligned}
    M_t&=\mathrm{sigmoid}\left( \beta \left( \xi _{flow}-E_{flow} \right) \right) \odot \mathrm{sigmoid}\left( \beta \left( \xi _{rgb}-E_{rgb} \right) \right) ,
    \end{aligned}
\end{equation}
\begin{equation}
    \label{eq: error}
    \begin{aligned}
    E_{flow}&=\mathrm{Norm}\left( F_{bwd,t}+\mathrm{Warp}_{t-1\rightarrow t}\left( F_{fwd,t-1} \right) \right) , \quad E_{rgb}=|I_t-\mathrm{Warp}_{t+1\rightarrow t}\left( I_{t+1} \right) |.
    \end{aligned}
\end{equation}

Here, $\beta$ is a constant scaling factor, $\xi _{flow}$ and $\xi _{rgb}$ are thresholds set from the statistics of error map $E_{flow}$ and $E_{rgb}$. This soft mask is also applied in the second stage of optimization. As shown in \cref{tab: ablation}, soft masking outperforms the hard one in both temporal consistency and prompt alignment.

\subsubsection{Stage II: Optimization over Unique Video Tensor}
\label{subsubsec: uvt}

In the second stage, we refine illumination and texture details. Compared with vanilla video, its canonical representation can incorporate spatial-temporal priors and facilitate consistency \cite{ouyang2024codef, shin2024enhancing}. But popular NeRF or 3DGS are too complex and costly for learning (cf. \cref{subsec: video editing}). Instead, we compress the video to a one-dimensional RGB vector of shape $(N,3)$, as shown in part (c) of \cref{fig: pipeline}. Specifically, we define a $d$-dimensional index $\kappa(x,y,t)$ for each pixel based on priors extracted from the source video. An example index could be $[22, 127, 0, 255]$, where the first element is the flow ID (pixels connected by the optical flow predicted by MemFlow share the same flow ID), and the rest are 8-bit quantized RGB values. It is also allowed to extend this 4-element index to more elements with voxel coordinate (from depth projection) or any other cues that indicate spatial–temporal similarity and locality. All pixels with identical $\kappa$ are gathered via averaging to form one element of the one-dimensional vector, where $N$ is the number of unique $\kappa$. Take the source video $\{I_t^{in}\}$ as an example, the \textbf{gathering} and \textbf{scattering} operations are formulated as

\begin{equation}
    \label{eq: uvt-org}
    \begin{aligned}
    \mathbf{U}\left( \kappa _n \right) =\mathrm{Avg}\left( \left\{ I_t^{in}\left( x,y \right) |\kappa (x,y,t)=\kappa _n \right\} \right) , \quad \mathbb{I}_t^{in}\left( x,y \right) =\mathbf{U}\left( \kappa (x,y,t) \right) , 
    \end{aligned}
\end{equation} 

where $\mathbf{U}$ is referred to as the \textbf{Unique Video Tensor (UVT)}. With an appropriate definition of $\kappa$, the scattered $\mathbb{I}_t^{in}\left( x,y \right)$ reconstructs the original $I_t^{in}\left( x,y \right)$ with minimal information loss, as validated in \cref{tab: uvt recon}. For relighting, the ideal edited video frames must preserve consistent motion and intrinsic image details with the source; thus, they share the same index tensor $\kappa$ for UVT representation. Accordingly, we compress the first-stage output $\tilde{I}_t\left( x,y \right)$ into $\mathbf{\tilde{U}}$ via \cref{eq: uvt-org}, which then serves as the primary optimization target. This formulation not only facilitates optimization but also naturally embeds spatial-temporal similarity priors (cf. \cref{subsec: ablation}). With CUDA parallelism, the gathering and scattering process can be performed instantly. The optimization of $\tilde{\mathbf{U}}$ is supervised by

% Though one could learn an auxiliary residual for the UVT, our empirical results indicate negligible performance improvement at the cost of increased complexity. Hence, it is omitted in our final implementation.

\begin{equation}
    \label{eq: uvt-loss}
    \begin{aligned}
    \mathcal{L} _{unique}=\lambda _{tv}\mathcal{L} _{tv}\left( \tilde{\mathbb{I}}_t \right) + \left( 1-\lambda _u \right) \mathcal{L} _{SSIM}\left( \tilde{\mathbb{I}}_t,\tilde{I}_t \right) + \\
    +\lambda _u\mathcal{L} _1\left( \tilde{\mathbb{I}}_t\odot M_t,\mathrm{Warp}_{t+1\rightarrow t}\left( \tilde{\mathbb{I}}_{t+1} \right) \odot M_t \right) ,
    \end{aligned}
\end{equation} 

where  $\tilde{\mathbb{I}}_t \left( x,y \right) =\tilde{\mathbf{U}}\left( \kappa (x,y,t) \right)$, and $\lambda _{tv}$ and $\lambda _u\in [0,1]$ balance the loss terms. The total variation loss $\mathcal{L} _{tv}$ suppresses noise. Notably, \cref{eq: uvt-loss} applies SSIM loss instead of photometric loss. This leaves space to fine-grained appearance and illumination adjustment without altering image structure. Finally, the optimized $\widehat{\mathbf{U}}$ is used to reconstruct $\widehat{\mathbb{I} }_t(x,y) $ according to \cref{eq: uvt-org} as the final output.

% We also allow learning a residual map $\tilde{r}_t\left( x,y \right)$ for better reconstruction:

% \begin{equation}
%     \label{eq: uvt-opt}
%     \begin{aligned}
%     \tilde{\mathbb{I}}_t\left( x,y \right) =\tilde{\mathbf{U}}\left( \kappa (x,y,t) \right) +\tilde{r}_t\left( x,y \right) .
%     \end{aligned}
% \end{equation} 

\section{Experiments}
\label{sec: experiments}

\subsection{Experiment Setting}
\label{subsec: exp setting}

\begin{table}
  \caption{Datasets \cite{MIFDB16, dosovitskiy2017carla, sun2020scalability, Dauner2024NEURIPS, contributors2025agibotworld, khazatsky2024droid, InteriorNet18, karnan2022scand} contained in established benchmark.  $N_{seq.}$ and $\bar{N}_{frames}$ denote number of sequence and average frames. C, F, D, S respectively denote RGB image, Optical Flow, Depth, Instance Segmentation. Notably, AgiBot here denotes AgiBot Digital World. Due to lacking of extrinsics, its depth is indeed not applicable. Only DRONE is self-collected data.} 
  \label{tab: dataset}
  \centering
  \resizebox{0.99\textwidth}{!}{%
  \begin{tabular}{cccccccccc}
    \toprule
    Datasets &SceneFlow &CARLA &Waymo &NavSim &AgiBot &DROID &InteriorNet &SCAND &DRONE \\
    \midrule
    Agent &Vehicle &Vehicle &Vehicle &Vehicle &Robot &Robot &Robot &Robot &Drone \\
    Synthetic &\checkmark &\checkmark & & &\checkmark & &\checkmark & &\\
    Modality &C,F,D,S &C,D,S &C &C &C &C &C,D,S &C &C \\
    $N_{seq.}$ &4 &8 &5 &5 &8 &12 &5 &6 &5 \\
    $\bar{N}_{frames}$ &300 &208 &198 &250 &305 &243 &300 &289 &213 \\
    Width &960 &960 &960 &960 &640 &960 &640 &960 &1280 \\
    Height &512 &536 &640 &536 &480 &536 &480 &536 &720 \\
    
    % Datasets & Scenario & & $N_{seq.}$ & $\bar{N}_{frames}$ & Resolution & Depth & Flow & Instance \\
    % SceneFlow \cite{MIFDB16} & Driving-Sim &4 &300 &$960\times512$ &\checkmark &\checkmark &\checkmark \\
    % CARLA \cite{dosovitskiy2017carla} & Driving-Sim & 8 &208 &$960\times536$ &\checkmark & &\checkmark  \\
    % Waymo \cite{sun2020scalability} & Driving-Real & 5 &198 &$960\times640$ & & & \\
    % NavSim \cite{Dauner2024NEURIPS} & Driving-Real & 5 &250 &$960\times536$ & & &  \\
    % AgiBot-Digital \cite{contributors2025agibotworld} & Manipulation-Sim & 6 &305 &$640\times480$ & & &  \\
    % DROID \cite{khazatsky2024droid} & Manipulation-Real &12 &243 &$960\times536$ \\
    % InteriorNet \cite{InteriorNet18} & Navigation-Sim &5 &300 &$640\times480$ &\checkmark & &\checkmark \\
    % SCAND \cite{karnan2022scand} & Navigation-Real &6 &289 &$960\times536$ \\
    % DRONE & Drone-Real &5 &213 &$1280\times720$\\
    \bottomrule
  \end{tabular}
  }
\end{table}

\textbf{Implementation Details.} Following IC-Light \cite{zhang2025scaling}, we apply $T=25$ sampling steps and a classifier-free guidance scale of 2.0. When inflated to video model with VidToMe \cite{li2024vidtome}, the local and global token merging ratios are 0.6 and 0.5, respectively, to accommodate high video dynamics. In our decayed multiaxis denoising strategy, the initial $\gamma_\tau$ is set to 0.2 and decays exponentially to 0.002 until the final sampling step. For the post-optimization stages, we use Adam \cite{kingma2017adammethodstochasticoptimization} as optimizer and run 35 epochs in the first stage and 70 in the second with a batch size of 16, ensuring fast yet sufficient convergence. $\kappa(x,y,t)$ mainly contains quantized RGB and estimated masked flow, and optionally depth if provided.  Emperically, the weighting coefficients $\lambda_{tv}$ is set to 0.01, $\lambda_e$ and $\lambda_u$ are set to 0.8. Following \cite{kerbl3Dgaussians}, the learning rate in the first stage decays from 0.01 to 0.001, while the second stage uses a fixed learning rate of 0.05. Additional details are included in the Appendix.

\textbf{Dataset.} To comprehensively evaluate the generation capability, we collect video clips with high motion dynamics and broad scene diversity. This \textbf{subjective} evaluation benchmark, as detailed in \cref{tab: dataset}, covers scenarios like autonomous driving, robot manipulation, and navigation, as well as drone flight. It includes data from synthetic and realistic environments under various weather conditions. Each clip is a long video with on average 256 frames, making it extremely challenging. To provide a more accurate and robust probe on the performance, we also conduct \textbf{objective} ground-truth-based evaluation on the Virtual KITTI 2 dataset \citep{cabon2020vkitti2}. We selected five scenes and relit them to match the illumination of morning, sunset, rain, overcast, and fog settings. Each sequence averages 281 frames at a resolution of 1248×384. To obtain edit prompts, we use some prompts from \cite{zhang2025scaling} and generate others using COSMOS \cite{agarwal2025cosmos}.

\begin{figure}
    \centering
    \includegraphics[width=0.99\linewidth]{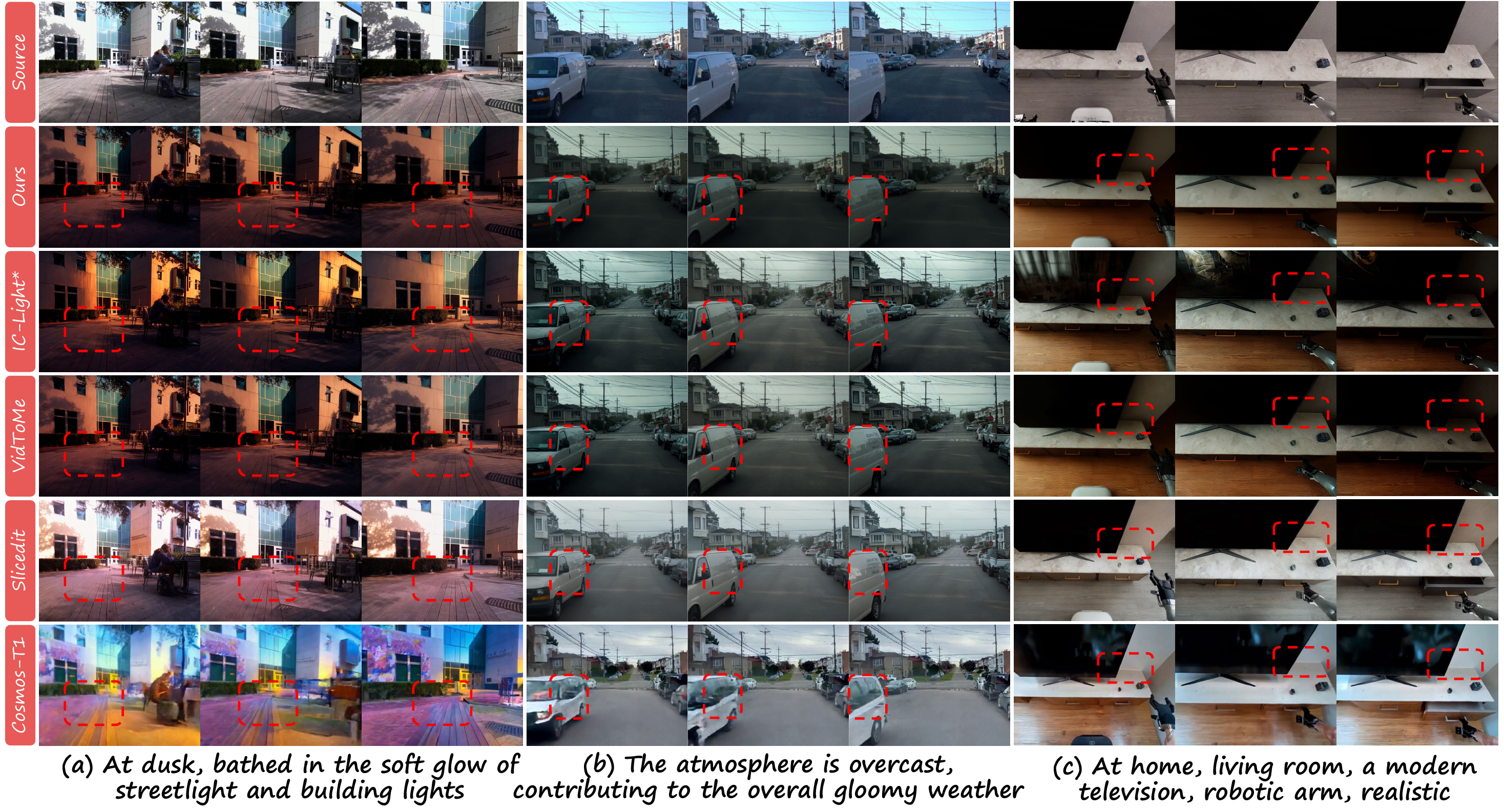}
    \caption{Qualitative comparison of results. The proposed TC-Light avoids unnatural relighting like Slicedit \cite{cohen2024slicedit} and COSMOS-Transfer1 \cite{alhaija2025cosmos} in (a) and blurring like \cite{alhaija2025cosmos} in (b), or inconsistent illumination like per-frame IC-Light \cite{zhang2025scaling} and VidToMe \cite{li2024vidtome} as highlighted by the \textcolor{red}{red} squares.}
    \label{fig: vis}
\end{figure}

\textbf{Metrics.} Following prior works \cite{qi2023fatezero, li2024vidtome, kara2024rave, wang2025videodirector}, we assess the relighting performance along following four dimensions: (i) \textit{Temporal consistency} is quantified via motion smoothness (\textbf{Motion-S}) \cite{huang2024vbench} and structural warping error (\textbf{Warp-SSIM}). Motion-S evaluates the continuity and physical plausibility of motion in the edited sequence, whereas Warp-SSIM computes the SSIM between a frame and its warped neighbors using flow from RAFT \cite{teed2020raft}. (ii) \textit{Textual alignment} is measured by average CLIP embedding similarity between the text prompt and all edited frames (\textbf{CLIP-T}). (iii) \textit{User preference} is evaluated by a study on 19 randomly selected videos and 65 valid submissions collected. Participants choose their preferred relighting results among our method and established baselines, from which we derive the Bradley–Terry preference rate (\textbf{User-PF}) \cite{bradley1952rank}. Additional details are included in the Appendix. (iv) \textit{Alignment with Groundtruth} is measured using \textbf{SSIM} and \textbf{LPIPS} \citep{kerbl3Dgaussians} with ground truth relighted results. These two metrics replace (ii) and (iii) on Virtual KITTI 2 \citep{cabon2020vkitti2} for a more accurate performance evaluation. (v) \textit{Computation efficiency} is reported in terms of runtime speed (FPS) and peak GPU memory consumption (VRAM) during editing. All experiments are conducted on a 40GB A100 GPU. Additionally, to appraise the reconstruction quality of UVT, we report average PSNR, SSIM, and LPIPS between the original and reconstructed frames.

\textbf{Baselines.} We benchmark our approach against several recent state-of-the-art techniques, whose code is publicly available at the time of writing. These include per-frame IC-Light (denoted as IC-Light*) and its video extensions, Light-A-Video \cite{zhou2025light} and RelightVid \cite{fang2025relightvid}. We also implement two IC-Light variants by incorporating leading zero-shot video editing methods: VidToMe \cite{li2024vidtome} and Slicedit \cite{cohen2024slicedit}. For fairness, we disable the image downsampling to $512\times512$ resolution before the diffusion step in Slicedit. In addition, we compare two advanced training-based methods—VideoDirector \cite{wang2025videodirector} and COSMOS-Transfer1 \cite{alhaija2025cosmos}. For the latter, due to out-of-memory (OOM) issues when applying full multimodal control on long videos, we employ only its edge branch, which offers a favorable balance between preserving image details and adhering to relighting prompts.

\subsection{Comparison with SOTA}
\label{subsec: comparison}

\begin{table}
  \caption{Comparison with existing methods. "OOM" here means the method is unable to finish the task due to an out-of-memory error. For a fair comparison, the base models of VidToMe and Slicedit are replaced with IC-Light here. \textbf{Ours-light} applies post-optimization to VidToMe, while \textbf{Ours-full} further introduces decayed multi-axis denoising. Experiments are conducted on 40G A100. The best and the second best of each metric are separately highlighted in \textcolor{red}{red} and \textcolor{blue}{blue}.}
  \label{tab: comparison}
  \centering
  \resizebox{0.99\textwidth}{!}{%
  \begin{tabular}{lccccccc}
    \toprule
    Method & Motion-S↑ & WarpSSIM↑ & CLIP-T↑ &User-PF↑ & FPS↑ &Time(s)↓ & VRAM(G)↓ \\
    \midrule
    IC-Light* \cite{zhang2025scaling} & 94.52\% &71.22 &\textcolor{red}{0.2743} &10.97\% &0.123 &2075 &16.49 \\
    VidToMe \cite{li2024vidtome} &95.38\% &73.69 &\textcolor{blue}{0.2731} &6.97\% &\textcolor{red}{0.409} &\textcolor{red}{626} &\textcolor{red}{11.65} \\
    Slicedit \cite{cohen2024slicedit} &96.48\% &85.37 &0.2653 &18.39\% &0.122 &2101 &17.87 \\
    VideoDirector \cite{wang2025videodirector} &OOM &OOM &OOM &OOM &OOM &OOM &OOM \\
    % FastBlend \cite{duan2023fastblend} &0.8985 &94.45 &0.2678 &19.70 &0.126 &2029 &11.65 \\
    \midrule
    Light-A-Video \cite{zhou2025light} &OOM &OOM &OOM &OOM &OOM &OOM &OOM \\
    RelightVid \cite{fang2025relightvid} &OOM &OOM &OOM &OOM &OOM &OOM &OOM \\
    Cosmos-T1 \cite{alhaija2025cosmos} &96.83\% &83.47 &0.2529 &16.06\% &0.101 &2543 &34.87 \\
    \midrule
    \textbf{Ours-light} &\textcolor{blue}{97.39\%} &\textcolor{blue}{88.53} &0.2700 &\textcolor{blue}{23.66\%} &\textcolor{blue}{0.359}  &\textcolor{blue}{771} &\textcolor{blue}{14.36} \\
    \textbf{Ours-full} &\textcolor{red}{97.80\%} &\textcolor{red}{91.75} &0.2679 &\textcolor{red}{23.96\%} &0.204 &1255 &14.37 \\
    \bottomrule
  \end{tabular}
  }
\end{table}

\begin{table}
  \caption{Comparison with existing methods on the Virtual KITTI 2 dataset \citep{cabon2020vkitti2}. The symbol definition aligns with \cref{tab: comparison}. Experiments are conducted on 40G A100. The best and the second best of each metric are separately highlighted in \textcolor{red}{red} and \textcolor{blue}{blue}.}
  \label{tab: comparison-kitti}
  \centering
  \resizebox{0.9\textwidth}{!}{%
  \begin{tabular}{lcccccc}
    \toprule
    Method & SSIM↑ & LPIPS↓ & Motion-S↑ &Warp-SSIM↑ &Time(s)↓ & VRAM(G)↓ \\
    \midrule
    IC-Light* \cite{zhang2025scaling} &0.5102 &0.4470 &95.23	&68.13	&1770 &\textcolor{blue}{10.25} \\
    VidToMe \cite{li2024vidtome} &0.5359 &0.4262 &95.95	&71.33	&\textcolor{red}{444} &\textcolor{red}{6.96} \\
    Slicedit \cite{cohen2024slicedit} &0.5080 &0.4237 &96.91 &80.74	&2346 &17.68 \\
    VideoDirector \cite{wang2025videodirector} &OOM &OOM &OOM &OOM &OOM &OOM \\
    % FastBlend \cite{duan2023fastblend} &0.8985 &94.45 &0.2678 &19.70 &0.126 &2029 &11.65 \\
    \midrule
    Light-A-Video \cite{zhou2025light} &OOM &OOM &OOM &OOM &OOM &OOM  \\
    RelightVid \cite{fang2025relightvid} &OOM &OOM &OOM &OOM &OOM &OOM \\
    Cosmos-T1 \cite{alhaija2025cosmos} &0.4833	&0.4841	&97.81	&84.35	&3314	&34.83 \\
    \midrule
    \textbf{Ours-light} &\textcolor{blue}{0.5855} &\textcolor{blue}{0.4026} &\textcolor{blue}{98.51} &\textcolor{blue}{90.94} &\textcolor{blue}{580} &15.21 \\
    \textbf{Ours-full} &\textcolor{red}{0.5910}	&\textcolor{red}{0.3971} &\textcolor{red}{98.62}	&\textcolor{red}{92.38}	&1002	&15.21 \\
    \bottomrule
  \end{tabular}
  }
\end{table}

Quantitative and qualitative comparisons with state-of-the-art methods are reported in \cref{tab: comparison}, \cref{tab: comparison-kitti}, and \cref{fig: vis}. The result indicates that per-frame relighting (IC-Light*) follows prompts well and produces physically plausible illumination, but the adapted illumination suffers from severe flicker, as shown in columns (a) and (b) of \cref{fig: vis}. IC-Light would even randomly hallucinate non-existent objects in textureless regions (cf. column (c) of \cref{fig: vis}), further degrading consistency. Extending IC-Light* with VidToMe \cite{li2024vidtome} yields modest gains in temporal coherence but dramatically lowers computation cost for long videos, so we adopt it as our primary baseline. Slicedit \cite{cohen2024slicedit} significantly suppresses flicker and hallucinations, yet its computation overhead exceeds that of IC-Light*. Besides, its output remains overly biased by the original appearance of the source video.  As a result, it produces unnatural relighting in many cases, as shown in column (a) of \cref{fig: vis}.

We also evaluated the T2V-model-based video editing approach \cite{wang2025videodirector} and concurrent video relighting techniques \cite{zhou2025light, fang2025relightvid}. Unfortunately, they all failed on long clips due to OOM errors caused by high computation resource demands. For the same reason, Cosmos-Transfer1 \cite{alhaija2025cosmos} can only operate in single-modality mode under GPU constraints, yet still requires over 30 GB GPU memory and more than 30 minutes per clip. Moreover, on video with high dynamics, it suffers from more severe blur and loss of details, as shown in columns (a) and (b) of \cref{fig: vis}. These failures are likely because Cosmos-Transfer1 is limited to the data domain of its training data, which contains less varied, moderately dynamic videos.

In contrast, our TC-Light first enables physically plausible relighting on long videos with high dynamics, while outperforming all baselines in temporal consistency and preference rate by a large margin, as shown in \cref{tab: comparison}. \cref{tab: comparison-kitti} also demonstrates that our model outperforms all baselines in perceptual and structural similarity with ground truth relighting results. Considering computation cost, the light version adds only 2.4 minutes and 2.7 GB of VRAM overhead compared to the VidToMe baseline (cf. \cref{tab: comparison}), while faithfully preserving object identity, albedo, and adherence to text prompts, as shown in \cref{fig: vis}. Incorporating our decayed multi-axis denoising further enhances temporal coherence, with a modest trade-off in efficiency and quality.  Limited by page, we provide additional visualization and performance of different scenarios types in the Appendix. And the video demos and comparison can be found on \href{https://dekuliutesla.github.io/tclight/}{our project page}.

\subsection{Ablation}
\label{subsec: ablation}

\begin{table}
  \caption{Ablation over module component. The experiments here are conducted on CARLA \cite{dosovitskiy2017carla} and the Interiornet \cite{InteriorNet18} subset, which both provide depth and instance mask as priors. There are 13 sequences in total and 254 frames on average, covering scenes of indoor and outdoor scenarios. The \colorbox{gray!20}{gray} row denotes modification that is aborted and not included in the following experiments.}
  \label{tab: ablation}
  \centering
  \tabcolsep=0.30cm
  \resizebox{0.99\textwidth}{!}{%
  \begin{tabular}{lcccccc}
    \toprule
    Method & Motion-S↑ & WarpSSIM↑ & CLIP-T↑ & FPS↑ &Time(s)↓ & VRAM(G)↓ \\
    \midrule
    Baseline &94.51\% &77.60 &0.2871 &0.693 &364 &10.63 \\
    +1st Stage &95.71\% &81.29 &0.2868 &0.651 &388 &11.33 \\
    % +2nd Stage(video, inst.) &0.8874 &92.39 &0.2355 &20.36 &0.482 &436 &12.90 \\
    % +2nd Stage(UVT, inst.) &0.8857 &92.79 &0.2341 &20.26 &0.481 &433 &11.57 \\
    +2nd Stage(video) &96.40\% &90.58 &0.2876 &0.552 &460 &13.53 \\
    +2nd Stage(UVT) &96.44\% &91.04 &0.2866 &0.563 &449 &11.81 \\
    % \rowcolor{gray!20} +residual & 96.42\% &91.03 &0.2866 &0.522 &487 &15.78 \\
    % +2nd Stage(video) &0.8859 &91.78 &0.2349 &20.35 &0.470 &450 &12.90\\
    % +2nd Stage(UVT) & \\0.8848 &92.30 &0.2344 &20.25 &0.473 &442 &11.57 \\
    % +1st Stage &0.8857 &92.79 &0.2341 &20.26 &0.481 &433 &11.57\\
    +soft mask &96.44\% &91.05 &0.2868 &0.559 &452 &11.81  \\
    \rowcolor{gray!20} from scratch(UVT) &96.30\% &90.65 &0.2866 &0.552 &458 &12.40 \\
    +depth &96.56\% &91.12 &0.2863 &0.569 &444 &11.57  \\ 
    \rowcolor{gray!20} +instance &96.50\% &91.01 &0.2851 &0.545 &462 &11.67 \\
    \midrule
    +multi-axis &98.41\% &95.52 &0.2813 &0.310 &805 &11.57\\
    +AIN &98.38\% &95.44 &0.2832 &0.310 &805 &11.57 \\
    +weight decay &97.75\% &93.74 &0.2865 &0.310 &805 &11.57 \\
    \bottomrule
  \end{tabular}
  }
\end{table}

\begin{table}
  \caption{Ablation over Unique Video Tensor (UVT). Here, \%$\mathrm{Cmpr}$ is the compression rate after applying UVT on the source video. The subscripts “f” and “f+d” indicate that, besides color cues, the UVT representation incorporates optical flow cues and both flow and depth cues, respectively.}
  \label{tab: uvt recon}
  \centering
  \resizebox{0.99\textwidth}{!}{%
  \begin{tabular}{l|cccc|cccc}
    \toprule
    Scene &\%$\mathrm{Cmpr}_{f}$↓ & $\mathrm{SSIM}_{f}$↑ & $\mathrm{PSNR}_{f}$↑ & $\mathrm{LPIPS}_{f}$↓ &\#$\mathrm{Cmpr}_{f+d}$↓ & $\mathrm{SSIM}_{f+d}$↑ & $\mathrm{PSNR}_{f+d}$↑ & $\mathrm{LPIPS}_{f+d}$↓\\
    \midrule
    CARLA &\%39.2 &0.9940 &50.71 &0.025 &\%29.2 &0.9925 &48.98 &0.028 \\
    InteriorNet &\%49.0 &0.9908 &46.17 &0.021 &\%12.8 &0.9755 &40.86 &0.047 \\
    \bottomrule
  \end{tabular}
  }
\end{table}

\begin{figure}
    \centering
    \includegraphics[width=0.99\linewidth]{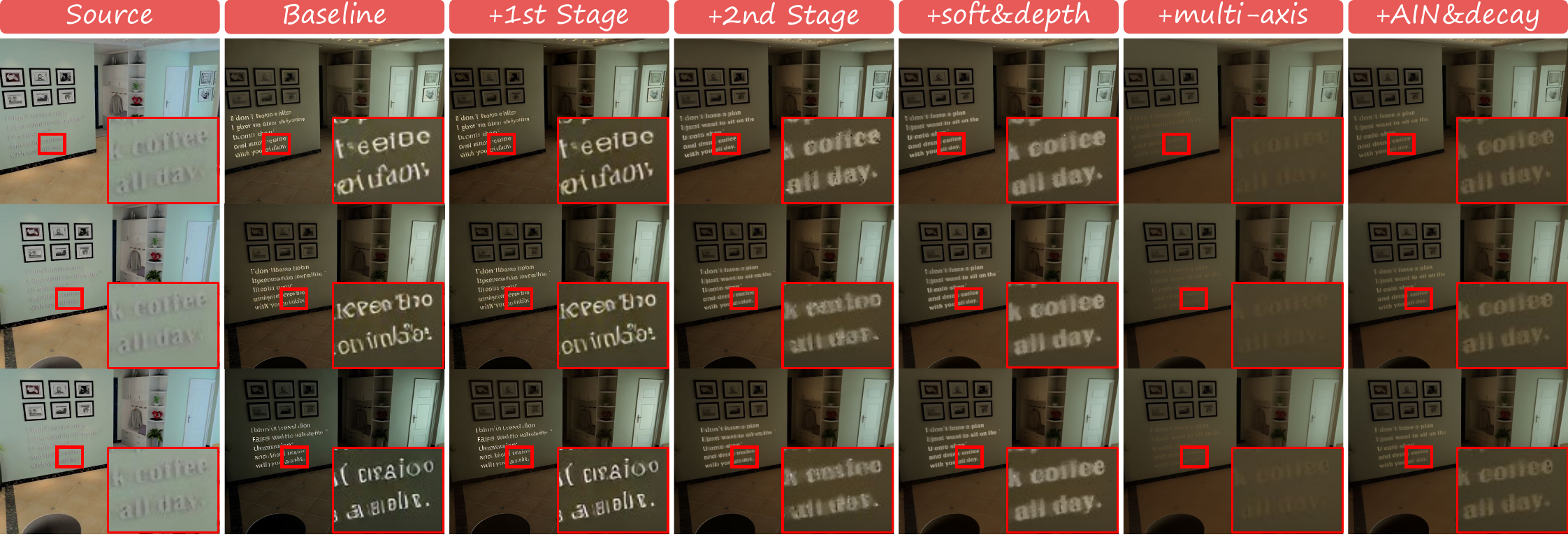}
    \caption{Ablation on main module components. The experiment is conducted on one sequence of the InteriorNet \cite{InteriorNet18} subset, where the text prompt is "This video showcases a modern interior space, which is dimly lit". The baseline here denotes VidToMe \cite{li2024vidtome} in \cref{tab: comparison}.}
    \label{fig: ablation}
\end{figure}

This section analyzes the contribution of each component in our model. The \textbf{first stage optimization}, as shown in \cref{tab: ablation} and \cref{fig: ablation}, markedly boosts consistency by aligning cross-frame exposure. The 6-7th rows of \cref{tab: ablation} also illustrate that, initializing UVT optimization from the first-stage results converges more efficiently than directly optimizing UVT for the same overall epochs. The \textbf{second stage optimization}, as shown in \cref{tab: ablation}, further reinforces temporal coherence. \cref{tab: uvt recon} confirms that UVT can compress the source video with near-zero loss, which underpins our design in \cref{subsubsec: uvt}. Using UVT as the second-stage target not only boosts consistency but also cuts computational overhead. Additionally, replacing a hard mask with a soft mask consistently improves both Warp-SSIM and CLIP-T metrics, demonstrating its importance. Incorporating the depth cues alongside UVT yields a more compact representation (also in \cref{tab: uvt recon}), which aids illumination alignment and release computation burden. In contrast, instance segmentation masks provide no clear benefit and are thus omitted from the final implementation.

% or learning auxiliary residual for UVT

For the \textbf{diffusion module}, multi-axis denoising notably enhances temporal consistency. However, it tends to inherit appearance distribution from the source video, causing drift from the target prompt and sometimes unnatural lighting, as shown in \cref{fig: ablation}. The introduced AIN and weight decay mitigate these issues, achieving a promising balance between consistency and faithful prompt alignment.

\subsection{Limitation and Discussion}
\label{subsec: limitation}

Despite achieving impressive results, our method is still limited by its base models. For instance, the current version of IC-Light \cite{zhang2025scaling} still struggles to relight hard shadows or make large modifications to low-light images. Similarly, since IC-Light is pretrained on 512 resolution and fine-tuned on 1024 resolution, our model struggles to preserve image details if the resolution is lower than 512.  For instance, downscaling the NavSim subset from 960×536 to 480×264 causes Warp-SSIM to drop from the average value 90.46 to 88.36, although CLIP-T remains at 0.304. Besides, since the optimization process relies on the optical flow estimation model, artifacts sometimes occur in textureless areas or under very fast motion, where the flow becomes unreliable. For instance, when downsampling videos from the NavSim subset by 4 times to simulate very fast motion, the CLIP-T and Motion Smoothness fluctuate less than 2\%, while Warp-SSIM declines by ~5\%. Furthermore, the temporal consistency loss has the tendency to smooth the texture of flickering areas, and therefore might sacrifice some details. It can be observed by comparing our model with IC-Light* in part (c) of \cref{fig: vis}. Though the proposed decayed multiaxis denoising alleviates the problem, developing a temporally more consistent and computationally more efficient denoising strategy is desired in future work.
\section{Conclusion}
% view-dependent effect; consistency limited by editing result and flow quality; detail loss and hallucination; hard shadow; long term drifting

In summary, we present TC-Light, a one-shot-tuned framework that delivers temporally consistent and physically plausible relighting on long, highly dynamic videos. The optimization-based illumination alignment provides a new paradigm for video relighting. Central to our approach is the Unique Video Tensor—an explicit, canonical, and differentiable video representation that enables highly efficient optimization. Over the established long video relighting benchmark, TC-Light achieves state-of-the-art performance in both consistency and efficiency, endowing it with value and potential for broader application areas such as sim2real and real-world video scaling in embodied AI training and validation pipelines.

\subsection*{Acknowledgments}
This work was supported in part by the National Natural Science Foundation of China (No. 62320106010, No. U21B2042) and the Guangdong Provincial Foshan Joint Funds (No. 2024A1515110065).

% \printbibliography{}
{
\small
\bibliographystyle{plain}
\bibliography{reference}

@article{bradley1952rank,
  title={Rank analysis of incomplete block designs: I. The method of paired comparisons},
  author={Bradley, Ralph Allan and Terry, Milton E},
  journal={Biometrika},
  volume={39},
  number={3/4},
  pages={324--345},
  year={1952},
  publisher={JSTOR}
}

@misc{kingma2017adammethodstochasticoptimization,
      title={Adam: A Method for Stochastic Optimization}, 
      author={Diederik P. Kingma and Jimmy Ba},
      year={2017},
      eprint={1412.6980},
      archivePrefix={arXiv},
      primaryClass={cs.LG},
      url={https://arxiv.org/abs/1412.6980}, 
}

@inproceedings{huang2024vbench,
  title={Vbench: Comprehensive benchmark suite for video generative models},
  author={Huang, Ziqi and He, Yinan and Yu, Jiashuo and Zhang, Fan and Si, Chenyang and Jiang, Yuming and Zhang, Yuanhan and Wu, Tianxing and Jin, Qingyang and Chanpaisit, Nattapol and others},
  booktitle={Proceedings of the IEEE/CVF Conference on Computer Vision and Pattern Recognition},
  pages={21807--21818},
  year={2024}
}

@inproceedings{teed2020raft,
  title={Raft: Recurrent all-pairs field transforms for optical flow},
  author={Teed, Zachary and Deng, Jia},
  booktitle={Computer Vision--ECCV 2020: 16th European Conference, Glasgow, UK, August 23--28, 2020, Proceedings, Part II 16},
  pages={402--419},
  year={2020},
  organization={Springer}
}

@inproceedings{dong2024memflow,
  title={Memflow: Optical flow estimation and prediction with memory},
  author={Dong, Qiaole and Fu, Yanwei},
  booktitle={Proceedings of the IEEE/CVF Conference on Computer Vision and Pattern Recognition},
  pages={19068--19078},
  year={2024}
}

@misc{cabon2020vkitti2,
  title={Virtual KITTI 2},
  author={Cabon, Yohann and Murray, Naila and Humenberger, Martin},
  year={2020},
  eprint={2001.10773},
  archivePrefix={arXiv},
  primaryClass={cs.CV}
}

@article{kerbl2024hierarchical,
  title={A hierarchical 3d gaussian representation for real-time rendering of very large datasets},
  author={Kerbl, Bernhard and Meuleman, Andreas and Kopanas, Georgios and Wimmer, Michael and Lanvin, Alexandre and Drettakis, George},
  journal={ACM Transactions on Graphics (TOG)},
  volume={43},
  number={4},
  pages={1--15},
  year={2024},
  publisher={ACM New York, NY, USA}
}

@inproceedings{2015-kingma,
  author = {Kingma, Diederik P. and Ba, Jimmy},
  booktitle = {ICLR (Poster)},
  editor = {Bengio, Yoshua and LeCun, Yann},
  ee = {http://arxiv.org/abs/1412.6980},
  title = {Adam: A Method for Stochastic Optimization.},
  url = {http://dblp.uni-trier.de/db/conf/iclr/iclr2015.html#KingmaB14},
  year = 2015
}

@article{mildenhall2021nerf,
  title={Nerf: Representing scenes as neural radiance fields for view synthesis},
  author={Mildenhall, Ben and Srinivasan, Pratul P and Tancik, Matthew and Barron, Jonathan T and Ramamoorthi, Ravi and Ng, Ren},
  journal={Communications of the ACM},
  volume={65},
  number={1},
  pages={99--106},
  year={2021},
  publisher={ACM New York, NY, USA}
}

@Article{kerbl3Dgaussians,
      author       = {Kerbl, Bernhard and Kopanas, Georgios and Leimk{\"u}hler, Thomas and Drettakis, George},
      title        = {3D Gaussian Splatting for Real-Time Radiance Field Rendering},
      journal      = {ACM Transactions on Graphics},
      number       = {4},
      volume       = {42},
      month        = {July},
      year         = {2023},
      url          = {https://repo-sam.inria.fr/fungraph/3d-gaussian-splatting/}
}

@InProceedings{MIFDB16,
  author    = "N. Mayer and E. Ilg and P. H{\"a}usser and P. Fischer and D. Cremers and A. Dosovitskiy and T. Brox",
  title     = "A Large Dataset to Train Convolutional Networks for Disparity, Optical Flow, and Scene Flow Estimation",
  booktitle = "IEEE International Conference on Computer Vision and Pattern Recognition (CVPR)",
  year      = "2016",
  note      = "arXiv:1512.02134",
  url       = "http://lmb.informatik.uni-freiburg.de/Publications/2016/MIFDB16"
}

@inproceedings{dosovitskiy2017carla,
  title={CARLA: An open urban driving simulator},
  author={Dosovitskiy, Alexey and Ros, German and Codevilla, Felipe and Lopez, Antonio and Koltun, Vladlen},
  booktitle={Conference on robot learning},
  pages={1--16},
  year={2017},
  organization={PMLR}
}

@inproceedings{Dauner2024NEURIPS,
	author = {Daniel Dauner and Marcel Hallgarten and Tianyu Li and Xinshuo Weng and Zhiyu Huang and Zetong Yang and Hongyang Li and Igor Gilitschenski and Boris Ivanovic and Marco Pavone and Andreas Geiger and Kashyap Chitta},
	title = {NAVSIM: Data-Driven Non-Reactive Autonomous Vehicle Simulation and Benchmarking},
	booktitle = {Advances in Neural Information Processing Systems (NeurIPS)},
	year = {2024},
}

@inproceedings{sun2020scalability,
  title={Scalability in perception for autonomous driving: Waymo open dataset},
  author={Sun, Pei and Kretzschmar, Henrik and Dotiwalla, Xerxes and Chouard, Aurelien and Patnaik, Vijaysai and Tsui, Paul and Guo, James and Zhou, Yin and Chai, Yuning and Caine, Benjamin and others},
  booktitle={Proceedings of the IEEE/CVF conference on computer vision and pattern recognition},
  pages={2446--2454},
  year={2020}
}

@article{contributors2025agibotworld,
  title={AgiBot World Colosseo: A Large-scale Manipulation Platform for Scalable and Intelligent Embodied Systems},
  author={AgiBot-World-Contributors and Bu, Qingwen and Cai, Jisong and Chen, Li and Cui, Xiuqi and Ding, Yan and Feng, Siyuan and Gao, Shenyuan and He, Xindong and Hu, Xuan and Huang, Xu and Jiang, Shu and Jiang, Yuxin and Jing, Cheng and Li, Hongyang and Li, Jialu and Liu, Chiming and Liu, Yi and Lu, Yuxiang and Luo, Jianlan and Luo, Ping and Mu, Yao and Niu, Yuehan and Pan, Yixuan and Pang, Jiangmiao and Qiao, Yu and Ren, Guanghui and Ruan, Cheng and Shan, Jiaqi and Shen, Yongjian and Shi, Chengshi and Shi, Mingkang and Shi, Modi and Sima, Chonghao and Song, Jianheng and Wang, Huijie and Wang, Wenhao and Wei, Dafeng and Xie, Chengen and Xu, Guo and Yan, Junchi and Yang, Cunbiao and Yang, Lei and Yang, Shukai and Yao, Maoqing and Zeng, Jia and Zhang, Chi and Zhang, Qinglin and Zhao, Bin and Zhao, Chengyue and Zhao, Jiaqi and Zhu, Jianchao},
  journal={arXiv preprint arXiv:2503.06669},
  year={2025}
}

@article{khazatsky2024droid,
    title   = {DROID: A Large-Scale In-The-Wild Robot Manipulation Dataset},
    author  = {Alexander Khazatsky and Karl Pertsch and Suraj Nair and Ashwin Balakrishna and Sudeep Dasari and Siddharth Karamcheti and Soroush Nasiriany and Mohan Kumar Srirama and Lawrence Yunliang Chen and Kirsty Ellis and Peter David Fagan and Joey Hejna and Masha Itkina and Marion Lepert and Yecheng Jason Ma and Patrick Tree Miller and Jimmy Wu and Suneel Belkhale and Shivin Dass and Huy Ha and Arhan Jain and Abraham Lee and Youngwoon Lee and Marius Memmel and Sungjae Park and Ilija Radosavovic and Kaiyuan Wang and Albert Zhan and Kevin Black and Cheng Chi and Kyle Beltran Hatch and Shan Lin and Jingpei Lu and Jean Mercat and Abdul Rehman and Pannag R Sanketi and Archit Sharma and Cody Simpson and Quan Vuong and Homer Rich Walke and Blake Wulfe and Ted Xiao and Jonathan Heewon Yang and Arefeh Yavary and Tony Z. Zhao and Christopher Agia and Rohan Baijal and Mateo Guaman Castro and Daphne Chen and Qiuyu Chen and Trinity Chung and Jaimyn Drake and Ethan Paul Foster and Jensen Gao and David Antonio Herrera and Minho Heo and Kyle Hsu and Jiaheng Hu and Donovon Jackson and Charlotte Le and Yunshuang Li and Kevin Lin and Roy Lin and Zehan Ma and Abhiram Maddukuri and Suvir Mirchandani and Daniel Morton and Tony Nguyen and Abigail O'Neill and Rosario Scalise and Derick Seale and Victor Son and Stephen Tian and Emi Tran and Andrew E. Wang and Yilin Wu and Annie Xie and Jingyun Yang and Patrick Yin and Yunchu Zhang and Osbert Bastani and Glen Berseth and Jeannette Bohg and Ken Goldberg and Abhinav Gupta and Abhishek Gupta and Dinesh Jayaraman and Joseph J Lim and Jitendra Malik and Roberto Martín-Martín and Subramanian Ramamoorthy and Dorsa Sadigh and Shuran Song and Jiajun Wu and Michael C. Yip and Yuke Zhu and Thomas Kollar and Sergey Levine and Chelsea Finn},
    year    = {2024},
}

@inproceedings { InteriorNet18,
      author = { Wenbin Li and Sajad Saeedi and John McCormac and Ronald Clark and 
                 Dimos Tzoumanikas and Qing Ye and Yuzhong Huang and Rui Tang and 
                 Stefan Leutenegger },
   booktitle = { British Machine Vision Conference (BMVC) },
       title = { InteriorNet: Mega-scale Multi-sensor Photo-realistic Indoor Scenes Dataset },
        year = { 2018 }
}

@article{karnan2022scand,
title = {Socially CompliAnt Navigation Dataset (SCAND): A Large-Scale Dataset Of Demonstrations For Social Navigation},
author = {Karnan, Haresh and Nair, Anirudh and Xiao, Xuesu and Warnell, Garrett and Pirk, S{\"o}ren and Toshev, Alexander and Hart, Justin and Biswas, Joydeep and Stone, Peter},
journal={IEEE Robotics and Automation Letters},
year = {2022},
organization = {IEEE}
}

@article{ho2020denoising,
  title={Denoising diffusion probabilistic models},
  author={Ho, Jonathan and Jain, Ajay and Abbeel, Pieter},
  journal={Advances in neural information processing systems},
  volume={33},
  pages={6840--6851},
  year={2020}
}

@inproceedings{rombach2022high,
  title={High-resolution image synthesis with latent diffusion models},
  author={Rombach, Robin and Blattmann, Andreas and Lorenz, Dominik and Esser, Patrick and Ommer, Bj{\"o}rn},
  booktitle={Proceedings of the IEEE/CVF conference on computer vision and pattern recognition},
  pages={10684--10695},
  year={2022}
}

@inproceedings{ronneberger2015u,
  title={U-net: Convolutional networks for biomedical image segmentation},
  author={Ronneberger, Olaf and Fischer, Philipp and Brox, Thomas},
  booktitle={Medical image computing and computer-assisted intervention--MICCAI 2015: 18th international conference, Munich, Germany, October 5-9, 2015, proceedings, part III 18},
  pages={234--241},
  year={2015},
  organization={Springer}
}

@article{saharia2022photorealistic,
  title={Photorealistic text-to-image diffusion models with deep language understanding},
  author={Saharia, Chitwan and Chan, William and Saxena, Saurabh and Li, Lala and Whang, Jay and Denton, Emily L and Ghasemipour, Kamyar and Gontijo Lopes, Raphael and Karagol Ayan, Burcu and Salimans, Tim and others},
  journal={Advances in neural information processing systems},
  volume={35},
  pages={36479--36494},
  year={2022}
}

@article{ramesh2022hierarchical,
  title={Hierarchical text-conditional image generation with clip latents},
  author={Ramesh, Aditya and Dhariwal, Prafulla and Nichol, Alex and Chu, Casey and Chen, Mark},
  journal={arXiv preprint arXiv:2204.06125},
  volume={1},
  number={2},
  pages={3},
  year={2022}
}

@article{agarwal2025cosmos,
  title={Cosmos world foundation model platform for physical ai},
  author={Agarwal, Niket and Ali, Arslan and Bala, Maciej and Balaji, Yogesh and Barker, Erik and Cai, Tiffany and Chattopadhyay, Prithvijit and Chen, Yongxin and Cui, Yin and Ding, Yifan and others},
  journal={arXiv preprint arXiv:2501.03575},
  year={2025}
}

@article{alhaija2025cosmos,
  title={Cosmos-Transfer1: Conditional World Generation with Adaptive Multimodal Control},
  author={Alhaija, Hassan Abu and Alvarez, Jose and Bala, Maciej and Cai, Tiffany and Cao, Tianshi and Cha, Liz and Chen, Joshua and Chen, Mike and Ferroni, Francesco and Fidler, Sanja and others},
  journal={arXiv preprint arXiv:2503.14492},
  year={2025}
}

@inproceedings{kara2024rave,
  title={Rave: Randomized noise shuffling for fast and consistent video editing with diffusion models},
  author={Kara, Ozgur and Kurtkaya, Bariscan and Yesiltepe, Hidir and Rehg, James M and Yanardag, Pinar},
  booktitle={Proceedings of the IEEE/CVF Conference on Computer Vision and Pattern Recognition},
  pages={6507--6516},
  year={2024}
}

@inproceedings{li2024vidtome,
  title={Vidtome: Video token merging for zero-shot video editing},
  author={Li, Xirui and Ma, Chao and Yang, Xiaokang and Yang, Ming-Hsuan},
  booktitle={Proceedings of the IEEE/CVF Conference on Computer Vision and Pattern Recognition},
  pages={7486--7495},
  year={2024}
}

@inproceedings{qi2023fatezero,
  title={Fatezero: Fusing attentions for zero-shot text-based video editing},
  author={Qi, Chenyang and Cun, Xiaodong and Zhang, Yong and Lei, Chenyang and Wang, Xintao and Shan, Ying and Chen, Qifeng},
  booktitle={Proceedings of the IEEE/CVF International Conference on Computer Vision},
  pages={15932--15942},
  year={2023}
}

@InProceedings{cohen2024slicedit,
    title={Slicedit: Zero-Shot Video Editing With Text-to-Image Diffusion Models Using Spatio-Temporal Slices},
    author={Cohen, Nathaniel and Kulikov, Vladimir and Kleiner, Matan and Huberman-Spiegelglas, Inbar and Michaeli, Tomer},
    booktitle={Proceedings of the 41st International Conference on Machine Learning},
    pages={9109--9137},
    year={2024},
    volume={235},
    series={Proceedings of Machine Learning Research},
    month={21--27 Jul},
    publisher={PMLR},
    url={https://proceedings.mlr.press/v235/cohen24a.html}
}

@inproceedings{huang2017arbitrary,
  title={Arbitrary style transfer in real-time with adaptive instance normalization},
  author={Huang, Xun and Belongie, Serge},
  booktitle={Proceedings of the IEEE international conference on computer vision},
  pages={1501--1510},
  year={2017}
}

@article{Richardt_Stoll_Dodgson_Seidel_Theobalt_2012,  
 title={Coherent Spatiotemporal Filtering, Upsampling and Rendering of RGBZ Videos}, 
 volume={31}, 
 url={https://doi.org/10.1111/j.1467-8659.2012.03003.x}, 
 DOI={10.1111/j.1467-8659.2012.03003.x}, 
 number={2pt1}, 
 journal={Computer Graphics Forum}, 
 author={Richardt, Christian and Stoll, Carsten and Dodgson, Neil A. and Seidel, Hans‐Peter and Theobalt, Christian}, 
 year={2012}, 
 month={May}, 
 pages={247–256}, 
 language={en-US} 
 }

@inproceedings{li2022physically,
  title={Physically-based editing of indoor scene lighting from a single image},
  author={Li, Zhengqin and Shi, Jia and Bi, Sai and Zhu, Rui and Sunkavalli, Kalyan and Ha{\v{s}}an, Milo{\v{s}} and Xu, Zexiang and Ramamoorthi, Ravi and Chandraker, Manmohan},
  booktitle={European Conference on Computer Vision},
  pages={555--572},
  year={2022},
  organization={Springer}
}

@inproceedings{jin2024neural_gaffer,
  title     = {Neural Gaffer: Relighting Any Object via Diffusion},
  author    = {Haian Jin and Yuan Li and Fujun Luan and Yuanbo Xiangli and Sai Bi and Kai Zhang and Zexiang Xu and Jin Sun and Noah Snavely},
  booktitle = {Advances in Neural Information Processing Systems},
  year      = {2024},
}

@inproceedings{ren2024relightful,
  title={Relightful harmonization: Lighting-aware portrait background replacement},
  author={Ren, Mengwei and Xiong, Wei and Yoon, Jae Shin and Shu, Zhixin and Zhang, Jianming and Jung, HyunJoon and Gerig, Guido and Zhang, He},
  booktitle={Proceedings of the IEEE/CVF Conference on Computer Vision and Pattern Recognition},
  pages={6452--6462},
  year={2024}
}

@inproceedings{kim2024switchlight,
  title={SwitchLight: Co-design of Physics-driven Architecture and Pre-training Framework for Human Portrait Relighting},
  author={Kim, Hoon and Jang, Minje and Yoon, Wonjun and Lee, Jisoo and Na, Donghyun and Woo, Sanghyun},
  booktitle={Proceedings of the IEEE/CVF Conference on Computer Vision and Pattern Recognition},
  pages={25096--25106},
  year={2024}
}

@inproceedings{wang2023sunstage,
  title={Sunstage: Portrait reconstruction and relighting using the sun as a light stage},
  author={Wang, Yifan and Holynski, Aleksander and Zhang, Xiuming and Zhang, Xuaner},
  booktitle={Proceedings of the IEEE/CVF Conference on Computer Vision and Pattern Recognition},
  pages={20792--20802},
  year={2023}
}

@inproceedings{
    zhang2025scaling,
    title={Scaling In-the-Wild Training for Diffusion-based Illumination Harmonization and Editing by Imposing Consistent Light Transport},
    author={Lvmin Zhang and Anyi Rao and Maneesh Agrawala},
    booktitle={The Thirteenth International Conference on Learning Representations},
    year={2025},
    url={https://openreview.net/forum?id=u1cQYxRI1H}
}

@inproceedings{cai2024real,
  title={Real-time 3d-aware portrait video relighting},
  author={Cai, Ziqi and Jiang, Kaiwen and Chen, Shu-Yu and Lai, Yu-Kun and Fu, Hongbo and Shi, Boxin and Gao, Lin},
  booktitle={Proceedings of the IEEE/CVF Conference on Computer Vision and Pattern Recognition},
  pages={6221--6231},
  year={2024}
}

@inproceedings{zhang2021neural,
  title={Neural video portrait relighting in real-time via consistency modeling},
  author={Zhang, Longwen and Zhang, Qixuan and Wu, Minye and Yu, Jingyi and Xu, Lan},
  booktitle={Proceedings of the IEEE/CVF international conference on computer vision},
  pages={802--812},
  year={2021}
}

@inproceedings{choi2024personalized,
  title={Personalized Video Relighting With an At-Home Light Stage},
  author={Choi, Jun Myeong and Christman, Max and Sengupta, Roni},
  booktitle={European Conference on Computer Vision},
  pages={394--410},
  year={2024},
  organization={Springer}
}

@misc{zhang2024lumisculptconsistencylightingcontrol,
      title={LumiSculpt: A Consistency Lighting Control Network for Video Generation}, 
      author={Yuxin Zhang and Dandan Zheng and Biao Gong and Jingdong Chen and Ming Yang and Weiming Dong and Changsheng Xu},
      year={2024},
      eprint={2410.22979},
      archivePrefix={arXiv},
      primaryClass={cs.CV},
      url={https://arxiv.org/abs/2410.22979}, 
}

@misc{bharadwaj2024genlitreformulatingsingleimagerelighting,
      title={GenLit: Reformulating Single-Image Relighting as Video Generation}, 
      author={Shrisha Bharadwaj and Haiwen Feng and Victoria Abrevaya and Michael J. Black},
      year={2024},
      eprint={2412.11224},
      archivePrefix={arXiv},
      primaryClass={cs.CV},
      url={https://arxiv.org/abs/2412.11224}, 
}

@inproceedings{kocsis2024lightit,
  title={Lightit: Illumination modeling and control for diffusion models},
  author={Kocsis, Peter and Philip, Julien and Sunkavalli, Kalyan and Nie{\ss}ner, Matthias and Hold-Geoffroy, Yannick},
  booktitle={Proceedings of the IEEE/CVF Conference on Computer Vision and Pattern Recognition},
  pages={9359--9369},
  year={2024}
}

@article{zhou2025light,
  title={Light-A-Video: Training-free Video Relighting via Progressive Light Fusion},
  author={Zhou, Yujie and Bu, Jiazi and Ling, Pengyang and Zhang, Pan and Wu, Tong and Huang, Qidong and Li, Jinsong and Dong, Xiaoyi and Zang, Yuhang and Cao, Yuhang and others},
  journal={arXiv preprint arXiv:2502.08590},
  year={2025}
}

@article{fang2025relightvid,
  title={RelightVid: Temporal-Consistent Diffusion Model for Video Relighting},
  author={Fang, Ye and Sun, Zeyi and Zhang, Shangzhan and Wu, Tong and Xu, Yinghao and Zhang, Pan and Wang, Jiaqi and Wetzstein, Gordon and Lin, Dahua},
  journal={arXiv preprint arXiv:2501.16330},
  year={2025}
}

@article{sun2019single,
  title={Single image portrait relighting.},
  author={Sun, Tiancheng and Barron, Jonathan T and Tsai, Yun-Ta and Xu, Zexiang and Yu, Xueming and Fyffe, Graham and Rhemann, Christoph and Busch, Jay and Debevec, Paul E and Ramamoorthi, Ravi},
  journal={ACM Trans. Graph.},
  volume={38},
  number={4},
  pages={79--1},
  year={2019}
}

@inproceedings{nestmeyer2020learning,
  title={Learning physics-guided face relighting under directional light},
  author={Nestmeyer, Thomas and Lalonde, Jean-Fran{\c{c}}ois and Matthews, Iain and Lehrmann, Andreas},
  booktitle={Proceedings of the IEEE/CVF Conference on Computer Vision and Pattern Recognition},
  pages={5124--5133},
  year={2020}
}

@article{pandey2021total,
  title={Total relighting: learning to relight portraits for background replacement.},
  author={Pandey, Rohit and Orts-Escolano, Sergio and Legendre, Chloe and Haene, Christian and Bouaziz, Sofien and Rhemann, Christoph and Debevec, Paul E and Fanello, Sean Ryan},
  journal={ACM Trans. Graph.},
  volume={40},
  number={4},
  pages={43--1},
  year={2021}
}

@inproceedings{das2021dsrn,
  title={Dsrn: an efficient deep network for image relighting},
  author={Das, Sourya Dipta and Shah, Nisarg A and Dutta, Saikat and Kumar, Himanshu},
  booktitle={2021 IEEE International Conference on Image Processing (ICIP)},
  pages={2788--2792},
  year={2021},
  organization={IEEE}
}

@misc{chen2023controlavideo,
        title={Control-A-Video: Controllable Text-to-Video Generation with Diffusion Models}, 
        author={Weifeng Chen and Jie Wu and Pan Xie and Hefeng Wu and Jiashi Li and Xin Xia and Xuefeng Xiao and Liang Lin},
        year={2023},
        eprint={2305.13840},
        archivePrefix={arXiv},
        primaryClass={cs.CV}
}

@misc{molad2023dreamixvideodiffusionmodels,
      title={Dreamix: Video Diffusion Models are General Video Editors}, 
      author={Eyal Molad and Eliahu Horwitz and Dani Valevski and Alex Rav Acha and Yossi Matias and Yael Pritch and Yaniv Leviathan and Yedid Hoshen},
      year={2023},
      eprint={2302.01329},
      archivePrefix={arXiv},
      primaryClass={cs.CV},
      url={https://arxiv.org/abs/2302.01329}, 
}

@misc{liew2023magicedit,
    author = {Liew, Jun Hao and Yan, Hanshu and Zhang, Jianfeng and Xu, Zhongcong and Feng, Jiashi},
    title = {MagicEdit: High-Fidelity and Temporally Coherent Video Editing},
    booktitle={arXiv},
    year = {2023}
}

@inproceedings{liang2024flowvid,
  title={Flowvid: Taming imperfect optical flows for consistent video-to-video synthesis},
  author={Liang, Feng and Wu, Bichen and Wang, Jialiang and Yu, Licheng and Li, Kunpeng and Zhao, Yinan and Misra, Ishan and Huang, Jia-Bin and Zhang, Peizhao and Vajda, Peter and others},
  booktitle={Proceedings of the IEEE/CVF Conference on Computer Vision and Pattern Recognition},
  pages={8207--8216},
  year={2024}
}

@inproceedings{feng2024ccedit,
  title={Ccedit: Creative and controllable video editing via diffusion models},
  author={Feng, Ruoyu and Weng, Wenming and Wang, Yanhui and Yuan, Yuhui and Bao, Jianmin and Luo, Chong and Chen, Zhibo and Guo, Baining},
  booktitle={Proceedings of the IEEE/CVF Conference on Computer Vision and Pattern Recognition},
  pages={6712--6722},
  year={2024}
}

@inproceedings{ma2024maskint,
  title={Maskint: Video editing via interpolative non-autoregressive masked transformers},
  author={Ma, Haoyu and Mahdizadehaghdam, Shahin and Wu, Bichen and Fan, Zhipeng and Gu, Yuchao and Zhao, Wenliang and Shapira, Lior and Xie, Xiaohui},
  booktitle={Proceedings of the IEEE/CVF Conference on Computer Vision and Pattern Recognition},
  pages={7403--7412},
  year={2024}
}

@inproceedings{qin2024instructvid2vid,
  title={Instructvid2vid: Controllable video editing with natural language instructions},
  author={Qin, Bosheng and Li, Juncheng and Tang, Siliang and Chua, Tat-Seng and Zhuang, Yueting},
  booktitle={2024 IEEE International Conference on Multimedia and Expo (ICME)},
  pages={1--6},
  year={2024},
  organization={IEEE}
}

@article{tokenflow2023,
    title = {TokenFlow: Consistent Diffusion Features for Consistent Video Editing},
    author = {Geyer, Michal and Bar-Tal, Omer and Bagon, Shai and Dekel, Tali},
    journal={arXiv preprint arxiv:2307.10373},
    year={2023}
}

@article{cong2023flatten,
  title={FLATTEN: optical FLow-guided ATTENtion for consistent text-to-video editing},
  author={Cong, Yuren and Xu, Mengmeng and Simon, Christian and Chen, Shoufa and Ren, Jiawei and Xie, Yanping and Perez-Rua, Juan-Manuel and Rosenhahn, Bodo and Xiang, Tao and He, Sen},
  journal={arXiv preprint arXiv:2310.05922},
  year={2023}
}

@inproceedings{chai2023stablevideo,
  title={Stablevideo: Text-driven consistency-aware diffusion video editing},
  author={Chai, Wenhao and Guo, Xun and Wang, Gaoang and Lu, Yan},
  booktitle={Proceedings of the IEEE/CVF International Conference on Computer Vision},
  pages={23040--23050},
  year={2023}
}

@inproceedings{ouyang2024codef,
  title={Codef: Content deformation fields for temporally consistent video processing},
  author={Ouyang, Hao and Wang, Qiuyu and Xiao, Yuxi and Bai, Qingyan and Zhang, Juntao and Zheng, Kecheng and Zhou, Xiaowei and Chen, Qifeng and Shen, Yujun},
  booktitle={Proceedings of the IEEE/CVF Conference on Computer Vision and Pattern Recognition},
  pages={8089--8099},
  year={2024}
}

@misc{shin2024enhancing,
      title={Enhancing Temporal Consistency in Video Editing by Reconstructing Videos with 3D Gaussian Splatting}, 
      author={Inkyu Shin and Qihang Yu and Xiaohui Shen and In So Kweon and Kuk-Jin Yoon and Liang-Chieh Chen},
      year={2024},
      eprint={2406.02541},
      archivePrefix={arXiv},
      primaryClass={cs.CV}
}

@inproceedings{wang2025videodirector,
  title={VideoDirector: Precise Video Editing via Text-to-Video Models},
  author={Wang, Yukun and Wang, Longguang and Ma, Zhiyuan and Hu, Qibin and Xu, Kai and Guo, Yulan},
  booktitle={Proceedings of the IEEE/CVF Conference on Computer Vision and Pattern Recognition},
  year={2025}
}

@inproceedings{zhang2022invrender,
  title={Modeling Indirect Illumination for Inverse Rendering},
  author={Zhang, Yuanqing and Sun, Jiaming and He, Xingyi and Fu, Huan and Jia, Rongfei and Zhou, Xiaowei},
  booktitle={CVPR},
  year={2022}
}

@article{zhang2021nerfactor,
  title={Nerfactor: Neural factorization of shape and reflectance under an unknown illumination},
  author={Zhang, Xiuming and Srinivasan, Pratul P and Deng, Boyang and Debevec, Paul and Freeman, William T and Barron, Jonathan T},
  journal={ACM Transactions on Graphics (ToG)},
  volume={40},
  number={6},
  pages={1--18},
  year={2021},
  publisher={ACM New York, NY, USA}
}
}

%%%%%%%%%%%%%%%%%%%%%%%%%%%%%%%%%%%%%%%%%%%%%%%%%%%%%%%%%%%%
\newpage
\appendix
\section*{Appendix}
\label{sec: appendix}

\section{Additional Experimental Results}

\begin{figure}
    \centering
    \includegraphics[width=0.99\linewidth]{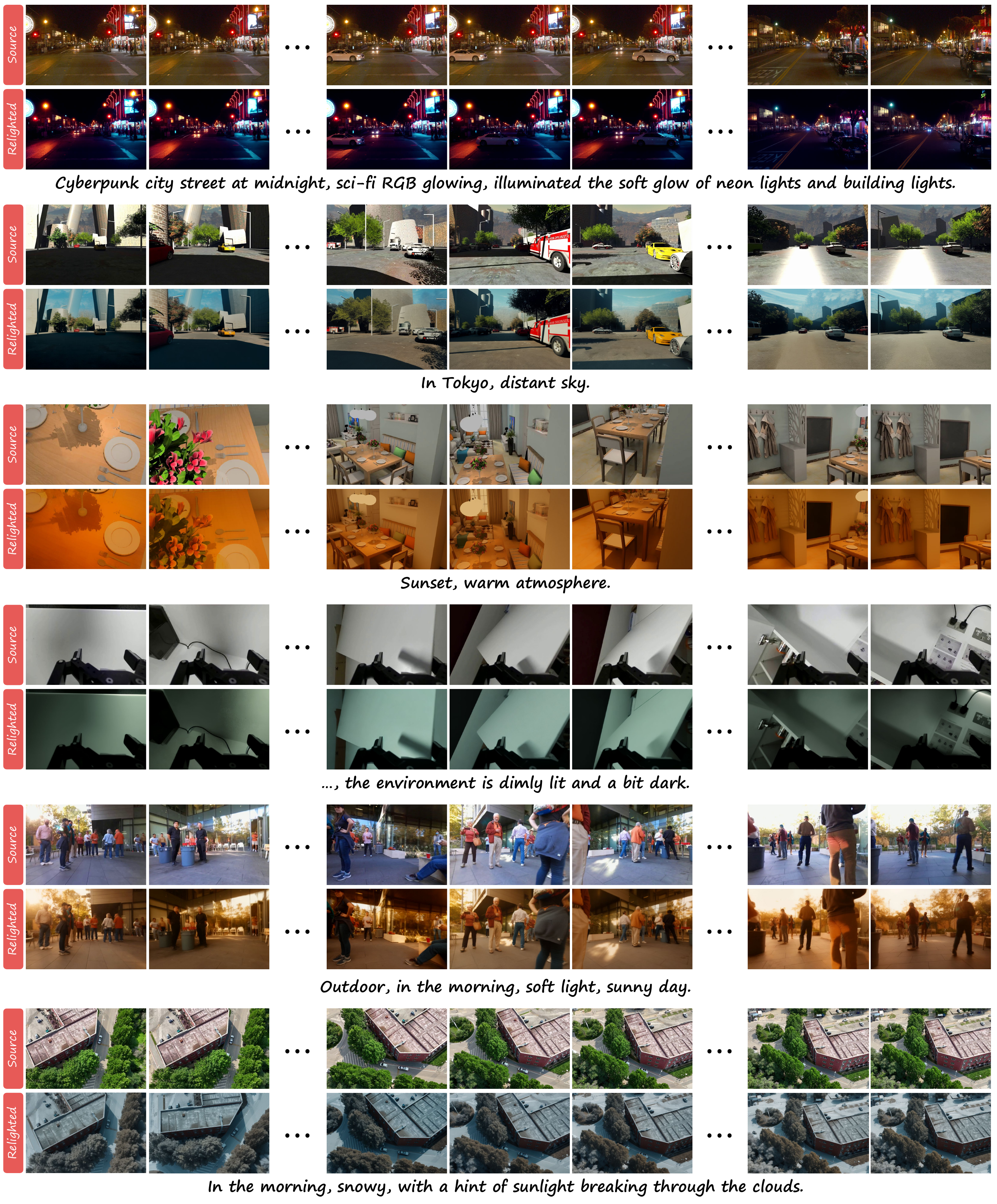}
    \caption{Qualitative results on additional long highly dynamic videos.}
    \label{fig: teaser_extra}
\end{figure}

\begin{figure}
    \centering
    \includegraphics[width=0.99\linewidth]{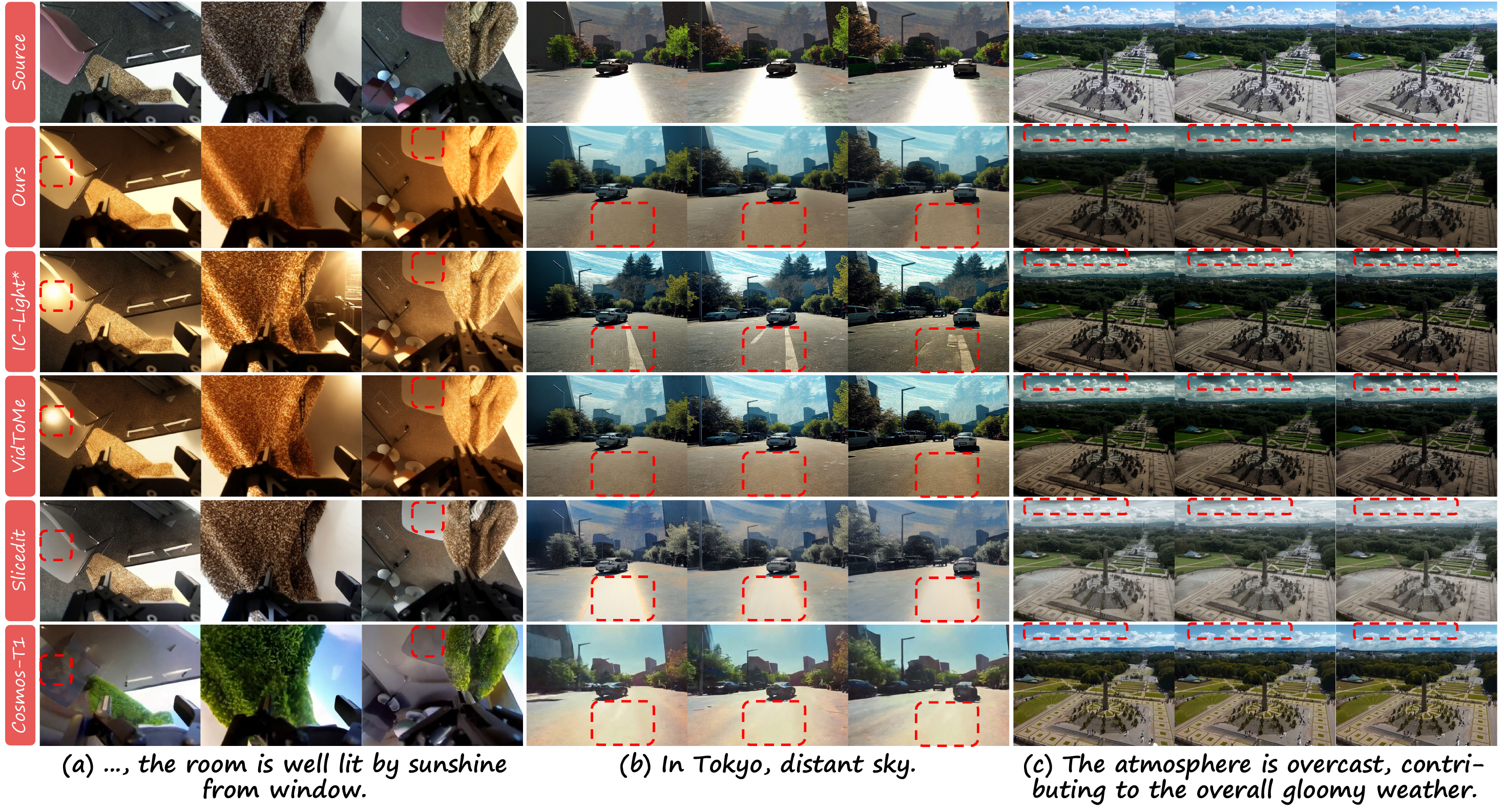}
    \caption{Additional qualitative comparison of results. The proposed TC-Light avoids unnatural relighting like Slicedit \cite{cohen2024slicedit} and COSMOS-Transfer1 \cite{alhaija2025cosmos} in (a) and (b), or temporal inconsistency like per-frame IC-Light \cite{zhang2025scaling} and VidToMe \cite{li2024vidtome} as highlighted by the \textcolor{red}{red} squares.}
    \label{fig: vis_extra}
\end{figure}

\begin{table}[h]
  \caption{Comparison on synthetic \cite{MIFDB16, dosovitskiy2017carla, contributors2025agibotworld, InteriorNet18} and realistic scenarios \cite{sun2020scalability, Dauner2024NEURIPS, khazatsky2024droid, karnan2022scand}. The average resolutions are respectively $794\times503$ and $960\times555$, while the frame numbers are 272 and 246. "OOM" here means the method is unable to finish the task due to an out-of-memory error. \textbf{Ours-light} applies post-optimization to VidToMe, while \textbf{Ours-full} further introduces decayed multi-axis denoising. The best and the second best of each metric are separately highlighted in \textcolor{red}{red} and \textcolor{blue}{blue}.}
  \label{tab: comparison-scenarios}
  \centering
  \tabcolsep=0.30cm
  \resizebox{0.99\textwidth}{!}{%
  \begin{tabular}{l|ccc|ccc}
    \toprule
    & &\textbf{Synthetic} & & &\textbf{Realistic} & \\
    Method & Motion-S↑ & WarpSSIM↑ & CLIP-T↑ &Motion-S↑ & WarpSSIM↑ & CLIP-T↑ \\
    \midrule
    IC-Light* \cite{zhang2025scaling} & 93.43\% &66.02 &\textcolor{red}{0.2779} &95.14\% &77.13 &\textcolor{red}{0.2837} \\
    VidToMe \cite{li2024vidtome} &94.61\% &69.45 &\textcolor{blue}{0.2776} &95.82\% &79.33 &0.2815 \\
    Slicedit \cite{cohen2024slicedit} &96.28\% &84.90 &0.2717 &96.38\% &88.89 &0.2715 \\
    VideoDirector \cite{wang2025videodirector} &OOM &OOM &OOM &OOM &OOM &OOM \\
    % FastBlend \cite{duan2023fastblend} &0.8985 &94.45 &0.2678 &19.70 &0.126 &2029 &11.65 \\
    \midrule
    Light-A-Video \cite{zhou2025light} &OOM &OOM &OOM &OOM &OOM &OOM \\
    RelightVid \cite{fang2025relightvid} &OOM &OOM &OOM &OOM &OOM &OOM \\
    Cosmos-T1 \cite{alhaija2025cosmos} &96.31\% &80.87 &0.2537 &96.78\% &83.57 &0.2659 \\
    \midrule
    \textbf{Ours-light} &\textcolor{blue}{97.02\%} &\textcolor{blue}{88.63} &0.2707 &\textcolor{blue}{97.46\%} &\textcolor{blue}{89.42} &\textcolor{blue}{0.2816} \\
    \textbf{Ours-full} &\textcolor{red}{97.36\%} &\textcolor{red}{91.07} &0.2695 &\textcolor{red}{97.90\%} &\textcolor{red}{92.67} &0.2792 \\
    \bottomrule
  \end{tabular}
  }
\end{table}

\cref{fig: teaser_extra} presents additional visualizations of our relighting results across a diverse range of scenarios. Whether under nighttime or daytime conditions, in outdoor or indoor environments, or from aerial or ground-level viewpoints, the proposed TC-Light method consistently produces temporally coherent and physically plausible illumination edits, demonstrating strong generalization capabilities. It is also worth noticing that the top row demonstrates our model’s ability to handle \textbf{spatially varying lighting}. In the middle three images, a white car drives from left to right. Initially, it is illuminated by orange street lamps, reflecting an orange hue. As it moves right, its rear remains orange-lit, while the front becomes blue due to a nearby advertising screen. Eventually, the car is fully bathed in blue light. This dynamic lighting response indicates our model can correctly handle spatially varying light.  \cref{fig: vis_extra} provides qualitative comparisons against state-of-the-art methods across additional scenarios. As shown, our model effectively adheres to textual instructions while generating relighting results that are both natural and temporally consistent.

We also provide corresponding quantitative evaluations on synthetic and real-world scenarios. As reported in \cref{tab: comparison-scenarios}, performance on real-world scenes consistently exceeds that on synthetic ones. This discrepancy likely arises from the training data of the video model Cosmos-Transfer1 \cite{alhaija2025cosmos} and the foundational image model IC-Light \cite{zhang2025scaling}, which are biased towards realistic scenes. Furthermore, the higher resolution and richer textures of real-world data mitigate hallucinations in textureless regions and help better preserve the intrinsic details of source frames for IC-Light. Such attributes are particularly critical for the consistency of methods with comparatively limited temporal modeling, namely, IC-Light* and VidToMe, which exhibit substantially higher Motion-S and WarpSSIM metrics on real-world videos than on synthetic ones. In contrast, our approach attains state-of-the-art temporal consistency across both scenario types while maintaining a favorable balance with prompt adherence.

% Notably, the reduction in CLIP-T is markedly smaller for real-world scenes compared to synthetic ones, possibly because flow estimation is less reliable in synthetic environments. The smoother, simpler textures therein degrade flow accuracy, lowering the quality of optimized outputs and thereby diminishing alignment with the text prompt.

\section{Details of Assets}
\label{sec: assets}

\begin{table}
  \caption{Licenses and video resolution of datasets \cite{MIFDB16, dosovitskiy2017carla, sun2020scalability, Dauner2024NEURIPS, contributors2025agibotworld, khazatsky2024droid, InteriorNet18, karnan2022scand} contained in established benchmark. Notably, AgiBot here denotes AgiBot Digital World. DRONE is our self-collected subset. Sceneflow has no license, but is only allowed for research purposes.} 
  \label{tab: dataset-resolution}
  \centering
  \resizebox{0.99\textwidth}{!}{%
  \begin{tabular}{cccccccccc}
    \toprule
    Datasets &SceneFlow &CARLA &Waymo &NavSim &AgiBot &DROID &InteriorNet &SCAND &DRONE \\
    \midrule
    Width &960 &960 &960 &960 &640 &960 &640 &960 &1280 \\
    Height &512 &536 &640 &536 &480 &536 &480 &536 &720 \\
    License & N/A &CC-BY & Custom\tablefootnote{\url{https://waymo.com/open/terms/}} &\makecell[c]{CC BY-\\NC-SA 4.0} &\makecell[c]{CC BY-\\NC-SA 4.0} &\makecell[c]{Apache\\-2.0} &Custom\tablefootnote{\url{https://interiornet.org/}} &CC0 1.0 &N/A \\
    \bottomrule
  \end{tabular}
  }
\end{table}

In \cref{tab: dataset-resolution}, we summarize the license and \textbf{resolution} for each subset. All source videos are resized and center-cropped to their designated resolutions. Considering the computation source limitation, we keep all frames if the sequence length is shorter than 300, and randomly sample 300 consecutive frames otherwise. Statistics are provided in Table 1 of the main paper. The DRONE subset includes three clips captured using our DJI Mini4 Pro and two additional clips obtained from DroneStock\footnote{\url{https://dronestock.com/}}, which are released under the CC0 1.0 License. For AgiBot Digital World \cite{contributors2025agibotworld}, where the robot's head moves in coordination with its body while performing tasks, relighting is performed from the head-mounted camera view. For each scene of DROID \cite{khazatsky2024droid}, we apply relighting to both the static side camera and the dynamic left wrist camera views. For Waymo \cite{sun2020scalability} and NavSim \cite{Dauner2024NEURIPS}, relighting is conducted using the front-facing camera view. \textbf{Domain balance} is maintained between synthetic and real environments (25 vs. 28 videos), also balanced within sub-domains: autonomous driving (12 synthetic, 10 real), robotic manipulation (8 synthetic, 12 real), indoor navigation (5 synthetic, 6 real). The aerial subset is excluded from balance due to limited long dynamic drone videos in the simulation environment and serves mainly for generalization validation

This paper also benefits from the code of IC-Light \cite{zhang2025scaling} (Apache-2.0 License), VidToMe \cite{li2024vidtome} (MIT License), Slicedit \cite{cohen2024slicedit} (MIT License), VideoDirector \cite{wang2025videodirector} (MIT License), Light-A-Video \cite{zhou2025light} (Apache-2.0 License), RelightVid \cite{fang2025relightvid} (CC BY-NC-SA 4.0 License), and Cosmos-Transfer1 \cite{alhaija2025cosmos} (Apache-2.0 License).

\section{Additional Implementation Details}
\label{sec: implementation}

For competing methods, we adopt the hyperparameters from their official implementations for  VideoDirector \cite{wang2025videodirector}, Light-A-Video \cite{zhou2025light}, RelightVid \cite{fang2025relightvid}, and Cosmos-T1 \cite{alhaija2025cosmos}. We replace base models of VidToMe \cite{li2024vidtome} and Slicedit \cite{cohen2024slicedit} with IC-Light \cite{zhang2025scaling}, and therefore we align their classifier-free guidance scale and diffusion sampling steps with those in \cite{zhang2025scaling}. Additionally, we set VidToMe’s local and global token‐merging ratios to 0.6 and 0.5, respectively, mirroring the setting of our approach. For Slicedit, we adjust the weighting factor $\gamma$ in Eq. (2) from the default 0.2 to 0.05 to better balance temporal coherence and instruction adherence. All other hyperparameters remain at their default values. The modified VidToMe serves as the baseline of our model design.

During implementation, each $\kappa(x,y,t)$ comprises three components: \textbf{(1)} a per-pixel flow ID, \textbf{(2)} a quantized RGB color, and, optionally, \textbf{(3)} a world-frame voxel coordinate. For (1), flow IDs are derived from the optical flow estimated by the state-of-the-art MemFlow method \cite{dong2024memflow} and the binary mask obtained by thresholding the soft mask in Eq. (6) of the main paper (values > 0.5 are set to 1; otherwise 0). In the initial frame, pixels receive unique flow IDs from $0$ to $HW-1$, where $H$ and $W$ denote image height and width. In subsequent frames, a pixel inherits the flow ID of its predecessor if connected by an unmasked flow; otherwise, it is assigned a new ID. This injects motion priors into the UVT representation. For \textbf{(2)}, we quantize RGB values to 7 bits, ensuring that all pixels sharing the same UVT element differ by less than $2/255$ in any channel. This constraint mitigates erroneous flows that escape the mask and reinforces representation in regions exhibiting view-dependent effects. For \textbf{(3)}, when per‐frame depth maps are available, they are reprojected into a point cloud using the camera intrinsics and extrinsics to determine world‐frame coordinates. This point cloud is then voxelized at a specified voxel size, and each pixel’s voxel coordinate is appended to $\kappa(x,y,t)$, yielding a more compact representation of static regions. For the CARLA \cite{dosovitskiy2017carla} and InteriorNet \cite{InteriorNet18} datasets, voxel sizes are set to 0.05 m and 0.02 m, respectively. Notably, dynamic objects at different timestamps may spatially overlap in 3D, but they remain distinguishable by their flow IDs and quantized RGB colors. Consequently, each object at each timestep is represented by a distinct set of UVT elements, while the $\mathcal{L}_{1}$ temporal consistency loss preserves object identity across frames.

\section{User Study}
\label{sec: user study}

\begin{figure}
    \centering
    \includegraphics[width=0.99\linewidth]{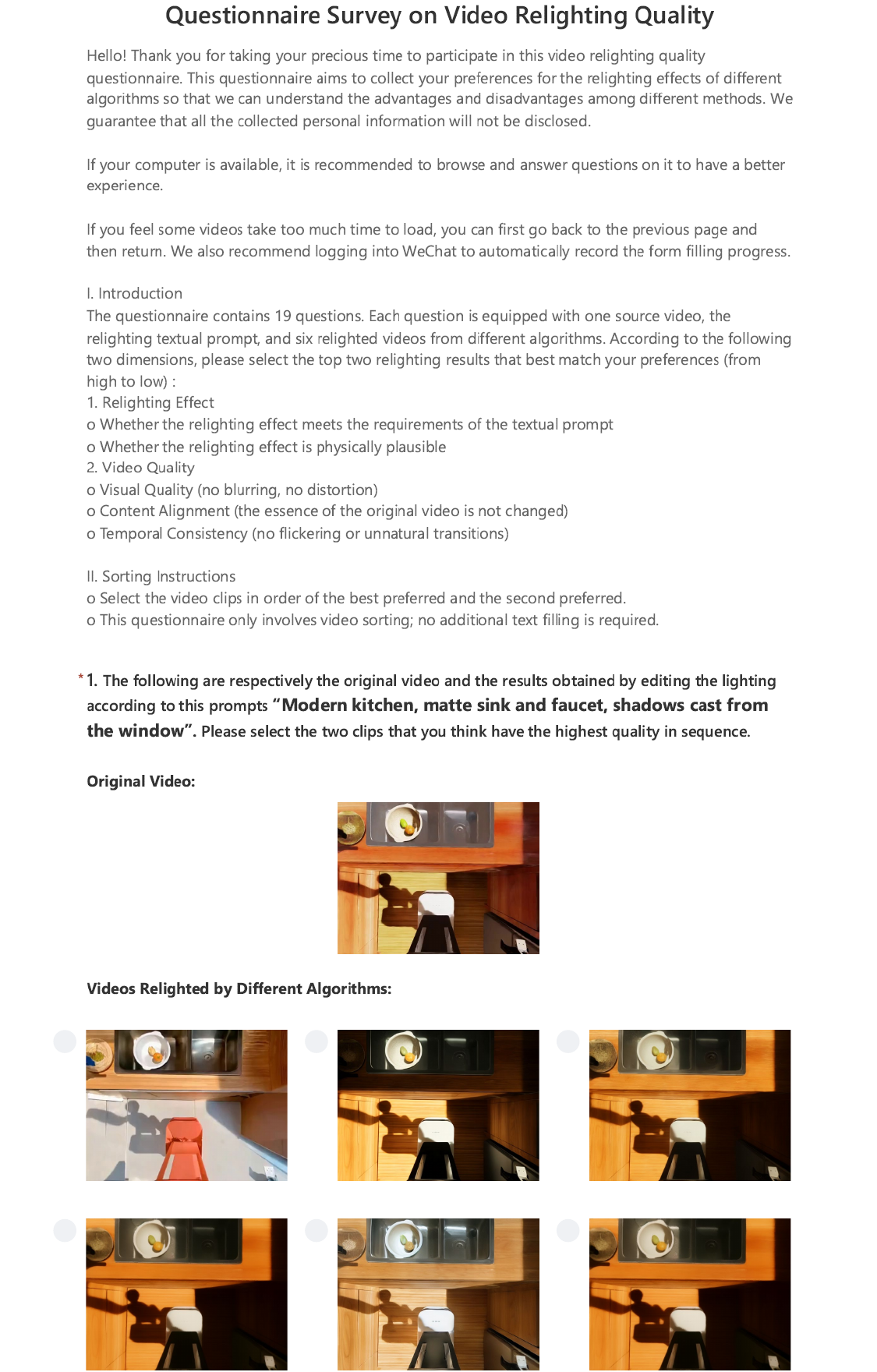}
    \caption{A screenshot of the user study.}
    \label{fig: questionnaire}
\end{figure}

We conducted an online user study with 78 anonymous participants, evaluating 19 randomly selected video–text pairs from our datasets. The compared methods were IC-Light* \cite{zhang2025scaling}, VidToMe \cite{li2024vidtome}, Slicedit \cite{cohen2024slicedit}, Cosmos-Transfer1 \cite{alhaija2025cosmos}, \textbf{Ours-light}, and \textbf{Ours-full}. A screenshot of the questionnaire interface is shown in \cref{fig: questionnaire}. For each question, methods were anonymized and relighted videos were presented in random order; participants selected the two most preferred results. In compliance with the NeurIPS Code of Ethics, each participant received a compensation of \$0.70. Besides, we ensured that all collected data remained confidential and was not disclosed to any institutions or individuals.

\begin{figure}
    \centering
    \includegraphics[width=0.85\linewidth]{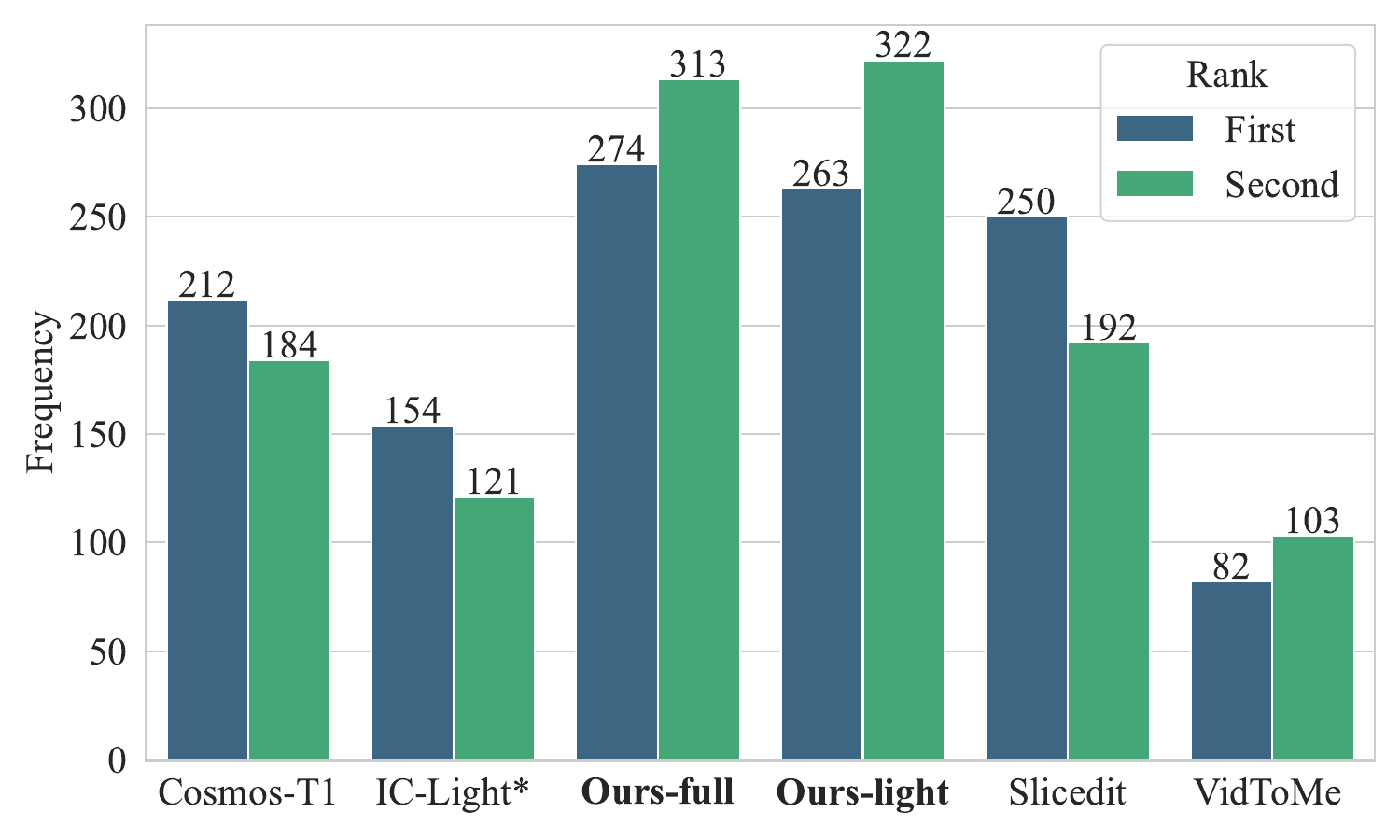}
    \caption{Results from user study with 65 valid submissions. The methods are arranged in alphabetical order. This figure reports the frequency that each method is chosen as the first- and second-most preferred video.}
    \label{fig: frequency}
\end{figure}

Since each video spanned 10–20 seconds, completing the questionnaire took on average 13.5 minutes. Submissions requiring less than four minutes were deemed unreliable and excluded, yielding 65 valid responses. \cref{fig: frequency} reports the frequency with which each method was chosen among the top two. Our full model achieved the highest preference rate, while the light variant ranked second. Although IC-Light* and VidToMe follow instructions well (cf. Tab. 2 of the main paper), their inferior temporal consistency make them much less preferred by users. Finally, we computed Bradley–Terry preference scores \cite{bradley1952rank} as a comprehensive metric of user preference, as presented in the Tab. 2 of the main paper.

\section{Physical Plausibility}
\label{sec: physical plausibility}

This section illustrates how our model maintains physical plausibility. The physical plausibility of our method is inherited from IC-Light \citep{zhang2025scaling}, which is pre-trained on high-quality Light Stage data and has learned a physically grounded relighting process. Our main contribution lies in improving temporal consistency without altering the illumination priors embedded in the base model.

As detailed in \cref{subsec: video model}, we introduce a video model inflation mechanism based on token merging/unmerging and decayed multi-axis denoising to enable temporal feature-level information exchange. Since these components do not alter the prior knowledge encoded in the base model, the distribution of edited illumination aligns with that of IC-Light while enhancing temporal consistency.

In Section \cref{subsec: post processing}, we propose a two-stage post-optimization strategy. The first stage smooths global exposure transitions using an appearance embedding, following practices in physically-based rendering methods like NeRF-W \citep{mildenhall2021nerf} and 3DGS \citep{kerbl3Dgaussians}. The second stage refines local fluctuations without altering the overall lighting. Thus, our final results maintain the physically plausible qualities of IC-Light, while significantly improving temporal coherence. As shown in \cref{fig: vis} of the main paper and the video results, our illumination remains qualitatively aligned with IC-Light and VidToMe, but with fewer artifacts and greater temporal stability.

Thanks to the strong priors of IC-Light, our method focuses on temporal coherence and computation efficiency. Compared to optimization-heavy approaches such as Nerfactor \citep{zhang2021nerfactor} and InvRender \citep{zhang2022invrender}, our post-optimization stage takes only around 2 minutes, with the entire pipeline completing in ~10 minutes—substantially faster than training NeRF or 3DGS models, as discussed in \cref{subsec: video editing} of the main paper

\section{Social Impact}
\label{sec: social impact}
\textbf{Positive Impacts.}
The proposed \textbf{TC-Light} framework for long video relighting stands to benefit a wide range of applications in both industry and research. First, by enabling consistent and physically plausible illumination editing at low computational cost, it can substantially lower the barrier to high‐quality visual content creation, empowering independent filmmakers, educators, and artists to produce compelling video narratives without access to specialized hardware. Second, the capability to scale illumination‐diverse training data through sim2real and real2real transfer can accelerate progress in embodied AI—robots and autonomous agents exposed to rich, temporally coherent visual environments may learn more robust perception and planning behaviors, thereby advancing safety and reliability in human–robot interaction. Finally, by fostering more efficient video synthesis pipelines, \textbf{TC-Light} may encourage energy‐aware design practices in large‐scale media processing systems, contributing to reduced resource consumption and attendant carbon emissions.

\textbf{Negative Impacts.}
Despite these benefits, improved video relighting carries potential risks if misused. Enhanced realism in dynamic relighting could facilitate the creation of deceptive multimedia, including deepfake videos that manipulate shadows and highlights to conceal tampering or impersonate individuals, thereby eroding trust in digital media. Moreover, large‐scale deployment of relighting tools raises privacy concerns: adversarial actors might relight surveillance footage to obscure identities or fabricate altered event sequences. To mitigate these harms, we advocate for gated access to pretrained models, integration of provenance metadata to flag relit content, and collaboration with platform providers to monitor and throttle suspicious bulk relighting requests.

%%%%%%%%%%%%%%%%%%%%%%%%%%%%%%%%%%%%%%%%%%%%%%%%%%%%%%%%%%%%

\newpage
\section*{NeurIPS Paper Checklist}
\begin{enumerate}

\item {\bf Claims}
    \item[] Question: Do the main claims made in the abstract and introduction accurately reflect the paper's contributions and scope?
    \item[] Answer: \answerYes{} % Replace by \answerYes{}, \answerNo{}, or \answerNA{}.
    \item[] Justification: The claims are supported by experiments in \cref{sec: experiments} both quantitatively and qualitatively. Since the dataset also covers the main application scenarios of embodied agents, we trust the potential of applying our work in embodied AI.

\item {\bf Limitations}
    \item[] Question: Does the paper discuss the limitations of the work performed by the authors?
    \item[] Answer: \answerYes{} % Replace by \answerYes{}, \answerNo{}, or \answerNA{}.
    \item[] Justification: Please refer to \cref{subsec: limitation}

\item {\bf Theory Assumptions and Proofs}
    \item[] Question: For each theoretical result, does the paper provide the full set of assumptions and a complete (and correct) proof?
    \item[] Answer: \answerNA{} % Replace by \answerYes{}, \answerNo{}, or \answerNA{}.
    \item[] Justification: No theoretical result is involved.

    \item {\bf Experimental Result Reproducibility}
    \item[] Question: Does the paper fully disclose all the information needed to reproduce the main experimental results of the paper to the extent that it affects the main claims and/or conclusions of the paper (regardless of whether the code and data are provided or not)?
    \item[] Answer: \answerYes{} % Replace by \answerYes{}, \answerNo{}, or \answerNA{}.
    \item[] Justification: The method is detailedly illustrated in \cref{sec: method}, and implementation details are provided in \cref{subsec: exp setting} and Appendix.

\item {\bf Open access to data and code}
    \item[] Question: Does the paper provide open access to the data and code, with sufficient instructions to faithfully reproduce the main experimental results, as described in supplemental material?
    \item[] Answer: \answerNo{} % Replace by \answerYes{}, \answerNo{}, or \answerNA{}.
    \item[] Justification: The full code and dataset is likely to be open-sourced upon acceptance. But the anonymous link to the partial dataset would be provided in the supplementary.

\item {\bf Experimental Setting/Details}
    \item[] Question: Does the paper specify all the training and test details (e.g., data splits, hyperparameters, how they were chosen, type of optimizer, etc.) necessary to understand the results?
    \item[] Answer: \answerYes{} % Replace by \answerYes{}, \answerNo{}, or \answerNA{}.
    \item[] Justification: Please refer to \cref{subsec: exp setting}.

\item {\bf Experiment Statistical Significance}
    \item[] Question: Does the paper report error bars suitably and correctly defined or other appropriate information about the statistical significance of the experiments?
    \item[] Answer: \answerNo{} % Replace by \answerYes{}, \answerNo{}, or \answerNA{}.
    \item[] Justification: Since both the quantitative experiments in comparison and ablation reports mean metrics over 10 video sequences, we trust that the fluctuation caused by noise is sufficiently suppressed.

\item {\bf Experiments Compute Resources}
    \item[] Question: For each experiment, does the paper provide sufficient information on the computer resources (type of compute workers, memory, time of execution) needed to reproduce the experiments?
    \item[] Answer: \answerYes{} % Replace by \answerYes{}, \answerNo{}, or \answerNA{}.
    \item[] Justification: All the quantitative results are accompanied by the execution time and memory cost. Since GPU is the main compute workers, we provide its details in \cref{subsec: exp setting}.
    
\item {\bf Code Of Ethics}
    \item[] Question: Does the research conducted in the paper conform, in every respect, with the NeurIPS Code of Ethics \url{https://neurips.cc/public/EthicsGuidelines}?
    \item[] Answer: \answerYes{} % Replace by \answerYes{}, \answerNo{}, or \answerNA{}.
    \item[] Justification: As far as we know, there is no break with NeurIPS Code of Ethics.

\item {\bf Broader Impacts}
    \item[] Question: Does the paper discuss both potential positive societal impacts and negative societal impacts of the work performed?
    \item[] Answer: \answerYes{} % Replace by \answerYes{}, \answerNo{}, or \answerNA{}.
    \item[] Justification: Due to the page limitation of the main paper, the discussion about positive and negative societal impacts of the work is included in the Appendix.
    
\item {\bf Safeguards}
    \item[] Question: Does the paper describe safeguards that have been put in place for responsible release of data or models that have a high risk for misuse (e.g., pretrained language models, image generators, or scraped datasets)?
    \item[] Answer: \answerNA{} % Replace by \answerYes{}, \answerNo{}, or \answerNA{}.
    \item[] Justification: This technique poses no such risks.

\item {\bf Licenses for existing assets}
    \item[] Question: Are the creators or original owners of assets (e.g., code, data, models), used in the paper, properly credited and are the license and terms of use explicitly mentioned and properly respected?
    \item[] Answer: \answerYes{} % Replace by \answerYes{}, \answerNo{}, or \answerNA{}.
    \item[] Justification: Due to the page limitation of the main paper, we list the license of used assets in the Appendix. 

\item {\bf New Assets}
    \item[] Question: Are new assets introduced in the paper well documented and is the documentation provided alongside the assets?
    \item[] Answer: \answerYes{} % Replace by \answerYes{}, \answerNo{}, or \answerNA{}.
    \item[] Justification: We include self-collected drone data and introduce it in \cref{subsec: exp setting}. Due to the page limitation of the main paper, we put details and the anonymized URL of the asset in the Appendix.

\item {\bf Crowdsourcing and Research with Human Subjects}
    \item[] Question: For crowdsourcing experiments and research with human subjects, does the paper include the full text of instructions given to participants and screenshots, if applicable, as well as details about compensation (if any)? 
    \item[] Answer: \answerYes{} % Replace by \answerYes{}, \answerNo{}, or \answerNA{}.
    \item[] Justification: This paper involves user study. Due to the page limitation of the main paper, the related details are provided in the Appendix.

\item {\bf Institutional Review Board (IRB) Approvals or Equivalent for Research with Human Subjects}
    \item[] Question: Does the paper describe potential risks incurred by study participants, whether such risks were disclosed to the subjects, and whether Institutional Review Board (IRB) approvals (or an equivalent approval/review based on the requirements of your country or institution) were obtained?
    \item[] Answer: \answerYes{} % Replace by \answerYes{}, \answerNo{}, or \answerNA{}.
    \item[] Justification: Yes, the potential risks are discussed in the Appendix and are disclosed to the subjects.

\end{enumerate}

\end{document}